%% file: main.tex
\documentclass[sigconf,9pt]{acmart}

\AtBeginDocument{%
  }
    
\acmConference[DAC '26]{Design Automation Conference}{July 26--29, 2026}{Long Beach, CA}
\setcopyright{none}
\renewcommand\footnotetextcopyrightpermission[1]{}
\usepackage{stfloats}
\usepackage{multirow}
\usepackage{bbding}
\usepackage{mathtools}
\usepackage{graphicx}
\usepackage{subcaption}
\usepackage{diagbox}
\usepackage{makecell}
\usepackage{textcomp}
\usepackage{pifont}
\usepackage{algorithm}
\usepackage{algpseudocode}
\algtext*{EndFor}
\algtext*{EndFunction}
\algtext*{Until}
\algtext*{EndIf}
\algtext*{EndWhile}
\usepackage{amsmath}
\usepackage{array}
\usepackage[referable]{threeparttablex}
\usepackage[T1]{fontenc}
\usepackage[utf8]{inputenc}
\usepackage{tablefootnote}
\usepackage{geometry}

\usepackage{listings}
\usepackage{xcolor}

\definecolor{codegreen}{rgb}{0,0.6,0}
\definecolor{codegray}{rgb}{0.5,0.5,0.5}
\definecolor{codepurple}{rgb}{0.58,0,0.82}
\definecolor{backcolour}{rgb}{0.95,0.95,0.92}

\lstdefinestyle{mystyle}{
  backgroundcolor=\color{backcolour},   commentstyle=\color{codegreen},
  keywordstyle=\color{magenta},
  numberstyle=\tiny\color{codegray},
  stringstyle=\color{codepurple},
  basicstyle=\ttfamily\footnotesize,
  breakatwhitespace=false,         
  breaklines=true,                 
  captionpos=b,                    
  keepspaces=true,                 
  numbers=left,                    
  numbersep=5pt,                  
  showspaces=false,                
  showstringspaces=false,
  showtabs=false,                  
  tabsize=2
}
\usepackage{colortbl}
\usepackage{dcolumn}

\definecolor{greenDeep}{RGB}{0,170,0}
\definecolor{greenSlightDeep}{RGB}{0,205,0}
\definecolor{greenShallow}{RGB}{0,255,0}
\definecolor{greenShallower}{RGB}{160,255,0}
\definecolor{orangeShallow}{RGB}{255,190,0}
\definecolor{orangeDeep}{RGB}{255,80,0}
\definecolor{orangeDeeper}{RGB}{255,40,0}
\definecolor{redDeep}{RGB}{255,0,0}

\definecolor{redLight}{RGB}{255,128,114}

\def\zz#1{%
\ifdim#1pt>4.9pt\cellcolor{greenDeep}\else
\ifdim#1pt>3.9pt\cellcolor{greenSlightDeep}\else
\ifdim#1pt>2.9pt\cellcolor{greenShallower}\else
\ifdim#1pt>2.9pt\cellcolor{yellow}\else
\ifdim#1pt>1.9pt\cellcolor{orangeShallow}\else
\ifdim#1pt>1.9pt\cellcolor{orange}\else
\ifdim#1pt>0.9pt\cellcolor{orange}\else
\ifdim#1pt>0.9pt\cellcolor{orangeDeep}\else
\cellcolor{orangeDeep}\fi\fi\fi\fi\fi\fi\fi\fi
#1}

\graphicspath{{./_fig/}}

\makeatletter
\renewcommand\footnoterule{%
  \kern-3\p@
  \hrule\@width0.4\columnwidth
  \kern2.6\p@}
  \makeatother

\newcommand{\yao}[1]{\textcolor{blue}{#1}} 
\newcommand{\gray}[1]{\textcolor{gray}{#1}} 
\newcommand{\ly}[1]{\textcolor{black}{#1}}

\definecolor{lightgreen}{RGB}{198, 224, 183}
\definecolor{lightred}{RGB}{240, 205, 176}

\usepackage[hang,flushmargin]{footmisc}

\begin{document}

%\title{OPTIC-RTL: Rethinking the Generation of PPA-Optimized RTL Code and A New Benchmark}
%\title{OPT-RTL: Rethinking the Optimization of RTL Code \\and A New Benchmark}
\title{A New Benchmark for the Appropriate Evaluation of \\RTL Code Optimization} 
% \author[1]{ }
\author{ 
\fontsize{10}{10}\selectfont 
Yao Lu, Shang Liu, Hangan Zhou, Wenji Fang, Qijun Zhang, Zhiyao Xie$^*$\\ 
\vspace{.05in}
% \fontsize{10}{10}\selectfont 
Hong Kong University of Science and Technology\\
% \vspace{.02in}
%\fontsize{10}{10}\selectfont 
\fontsize{8}{8}\selectfont ($^*$Corresponding Author: eezhiyao@ust.hk)}

%  \vspace{.05in}
% \\

\input{_txt/abstract}
\maketitle
\pagestyle{plain}

\input{_txt/1-intro}

\input{_txt/2-discussion}

\input{_txt/3-dataset}

\input{_txt/4-experiments}

\input{_txt/5-conclusion}

\bibliographystyle{ACM-Reference-Format}
\bibliography{iclr2026_conference}
\newpage
\onecolumn
\appendix
\clearpage
\input{_txt/appendix}
\end{document}

%% file: _txt/abstract.tex
\begin{abstract}
The rapid progress of artificial intelligence increasingly relies on efficient integrated circuit (IC) design. Recent studies have explored the use of large language models (LLMs) for generating Register Transfer Level (RTL) code, but existing benchmarks mainly evaluate syntactic correctness rather than optimization quality in terms of power, performance, and area (PPA). This work introduces RTL-OPT, a benchmark for assessing the capability of LLMs in RTL optimization. RTL-OPT contains 36 handcrafted digital designs that cover diverse implementation categories including combinational logic, pipelined datapaths, finite state machines, and memory interfaces. Each task provides a pair of RTL codes, a suboptimal version and a human-optimized reference that reflects industry-proven optimization patterns not captured by conventional synthesis tools. Furthermore, RTL-OPT integrates an automated evaluation framework to verify functional correctness and quantify PPA improvements, enabling standardized and meaningful assessment of generative models for hardware design optimization\footnote{RTL-OPT is available at \url{https://anonymous.4open.science/r/RTL-OPT-20C5}. Detailed experimental results and comprehensive evaluation of RTL-OPT are also included.}.

\end{abstract}

%% file: _txt/1-intro.tex
\section{Introduction}\label{sec:intro}

%The rapid evolution of artificial intelligence (AI) is increasingly intertwined with advancements in integrated circuit (IC) design, particularly as large language models (LLMs) begin to play a pivotal role in generating Register-Transfer Level (RTL) code. 

% many research works have started exploring 

%\yao{[1st paragraph: Introduce RTL code generaiton (without PPA optimization).]}

% The rapid advancements of AI rely on the support of integrated circuits (ICs), which are increasingly complex and difficult to optimize. 
In recent years, the adoption of large language models (LLMs) in the agile design of ICs has emerged as a promising research direction~\cite{fang2025survey}. Especially, many recent works \cite{liu2024rtlcoder, ho2024verilogcoder, liu2023chipnemo, pei2024betterv, fu2023gpt4aigchip, chang2023chipgpt, thakur2023autochip, cui2024origen, liu2024craftrtl, liu2025deeprtl, zhao2024codev} develop customized LLMs to directly generate IC designs in the format of Register-Transfer Level (RTL) code, such as Verilog.
% or VHDL. 

\textbf{Benchmarking RTL Code Generation.} The RTL design is the starting point of digital IC design implementation and requires significant human efforts and expertise. LLM-assisted RTL code generation techniques \cite{liu2024rtlcoder, ho2024verilogcoder, liu2025deeprtl, zhao2024codev} aim to relieve engineers from the tedious RTL coding process. 
To enable a fair comparison among different LLMs' capabilities in RTL generation, high-quality benchmarks become necessary. 
Representative benchmarks on RTL code generation include VerilogEval \cite{liu2023verilogeval} and RTLLM \cite{lu2024rtllm}, VerilogEval v2~\cite{pinckney2024revisiting}, RTLLM 2.0~\cite{liu2024openllm}, CVDP~\cite{pinckney2025comprehensive}, and others~\cite{delorenzo2024creativeval, allam2024rtl}.

\textbf{Difficulty in Evaluating RTL Optimization.} However, the aforementioned benchmarks primarily focus on the \emph{correctness} of RTL code generation, without explicitly evaluating the \emph{optimization of IC design's ultimate qualities} in terms of power, performance, and area (PPA). 
% Such PPA quality is a unique property of hardware RTL code, in comparison with software code. In hardware design, 
The RTL code will be synthesized into ultimate circuit implementations using synthesis tools, which will apply extensive logic optimizations when converting RTL code into implementations. Thus, PPA results depend on both the RTL code quality and the downstream synthesis process. As we will point out in this paper, this tight interplay makes benchmarking RTL optimization particularly challenging, sometimes even misleading.

%synthesis is similar to a compiler in software: it converts RTL code to circuit implementation while also applying extensive logic optimizations. Thus, PPA results depend on both the RTL code quality and the synthesis tool. As we will point out in this paper, this tight interplay makes benchmarking RTL optimization particularly challenging.

%\yao{[2nd paragraph: Introduce RTL code generation (with PPA optimization).]} 
%but also better ultimate chip quality. 

\begin{figure}[!t]
\centering
\begin{subfigure}{0.99\linewidth}
  \centering
  \includegraphics[width=\linewidth]{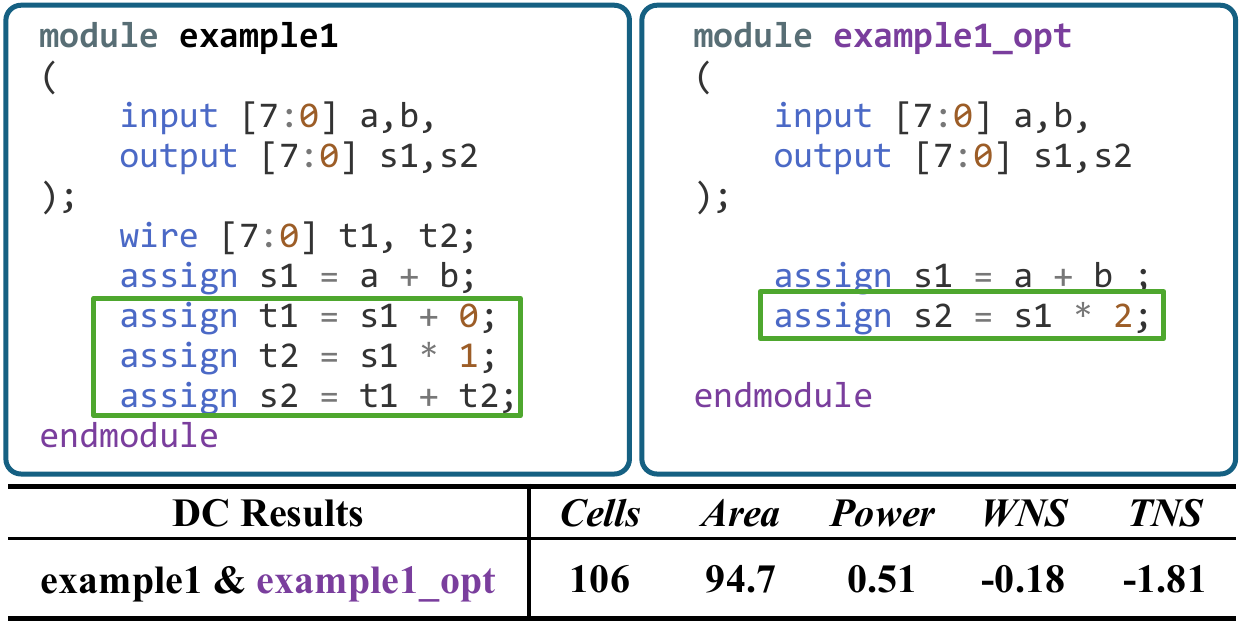}
  % \vspace{-.15in}
  \caption{An example of overly contrived \emph{suboptimal} and \emph{optimized} RTL code pair in existing benchmark~\cite{yao2024rtlrewriter}.}
  \label{fig:intro_a}
\end{subfigure}
\\ \vspace{0.1in}
\begin{subfigure}{0.99\linewidth}
  \centering
  \includegraphics[width=\linewidth]{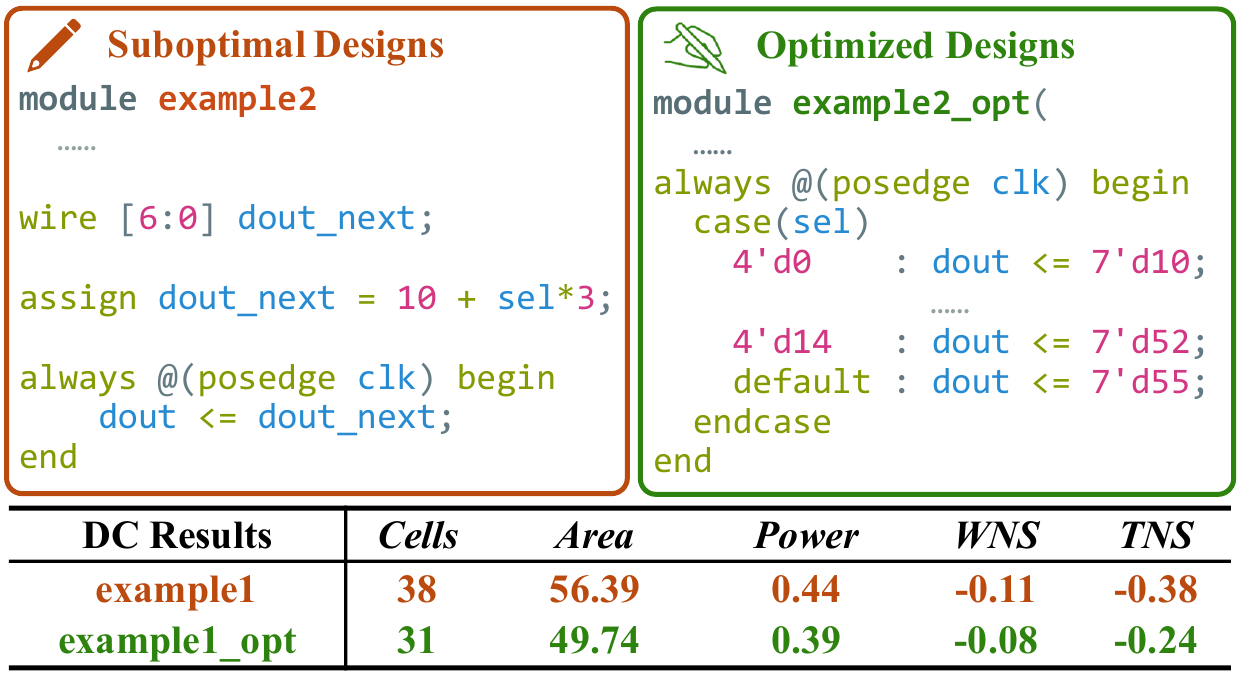}
  \caption{An example of \emph{suboptimal} and \emph{optimized} RTL code in our benchmark RTL-OPT, reflecting realistic optimized opportunities.}
  \label{fig:intro_b}
\end{subfigure}
\vspace{-.05in}
\caption{Comparison of examples from the few existing RTL optimization works~\cite{yao2024rtlrewriter} and our proposed benchmark, which captures realistic RTL optimization opportunities. The suboptimal design is the input for LLMs to improve and the optimized design is a golden reference.}
\label{fig:intro}
\vspace{-.2in}
\end{figure}

\textbf{Benchmarking RTL Code Optimization.} 
Constructing a high-quality benchmark for RTL optimization is inherently challenging due to the severe scarcity of open-source circuit designs, which are valuable IPs for semiconductor companies. 
Most recently, some LLM works~\cite{yao2024rtlrewriter, wang2025symrtlo, xu2025rethinkingllmbasedrtlcode} start to target generating more \emph{optimized} RTL code, which is expected to yield better ultimate chip quality in PPA. 
\ly{These works are all evaluated on the only relevant benchmark~\cite{yao2024rtlrewriter}, which provides suboptimal RTL codes for LLMs to improve.}
However, our study indicates that this benchmark~\cite{yao2024rtlrewriter} falls short in several aspects:  
1) \textbf{Unrealistic designs:} many suboptimal RTL codes in this benchmark are overly contrived and fail to capture real inefficiencies in practice;  
2) \textbf{Oversimplified synthesis setup:} reliance on weak synthesis tools such as Yosys~\cite{wolf2013yosys} leads to results that are sensitive to superficial RTL code changes and poorly aligned with industrial-grade flows; 
3) \textbf{Insufficient evaluation:} its assessments \ly{focus only on area-related metrics, while neglecting power and timing. Such evaluation metric neglects the ubiquitous trade-offs in a typical IC design process.}

In this work, we first inspect the existing works on RTL optimization and rethink a key question: \textbf{how to benchmark the optimization of RTL code appropriately?}\footnote{Please note that the pioneering work~\cite{yao2024rtlrewriter} did not claim their designs as a standard benchmark, and we appreciate their open-source contributions.}
We carefully inspect existing works and downstream synthesis flows.
This study reveals that evaluating RTL optimization is non-trivial and may easily lead to misleading conclusions. Specifically, whether one RTL code is superior (i.e., more optimized) to the other strongly depends on the synthesis tool and setup. 
Many ``optimized'' RTL codes indicated by the prior work~\cite{yao2024rtlrewriter} turn out to be the same or even worse than their ``suboptimal'' RTL counterparts when different, typically more advanced, synthesis options are adopted. 

% \looseness=-1

Figure~\ref{fig:intro_a} highlight a common flaw in \cite{yao2024rtlrewriter} designs: its RTL pairs are based on unrealistic transformations that do not address true optimization challenges in hardware design. \ly{These contrived suboptimal examples exhibit unnecessary inefficiencies, such as redundant computations and superfluous arithmetic operations, which are unlikely to occur in practice. Synthesis tools can easily optimize these contrived patterns, leading to evaluations that may overstate the effectiveness of LLMs in improving RTL quality.} 
The optimized version (\texttt{example1\_opt}) implements the logic directly by computing \texttt{s1 = a + b} and then deriving the output as \texttt{s2 = s1 * 2}. In contrast, the suboptimal version (\texttt{example1}) introduces contrived and unnecessary steps: it first computes \texttt{s1 = a + b}, then redundantly adds 0 and multiplies by 1 to produce intermediate wires \texttt{t1} and \texttt{t2}, before summing them into \texttt{s2}. Such constructions are unnatural and would rarely appear in practical RTL coding, making the benchmark examples unrealistic. Figure~\ref{fig:intro_b} shows an example from RTL-OPT, the detailed analysis are in Section~\ref{sec:intro_b}.

\begin{figure*}[!t]
\vspace{-.2in}
\includegraphics[width=1\textwidth]{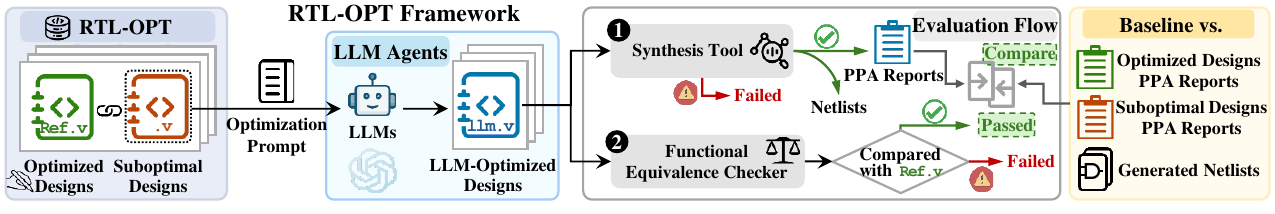}
\vspace{-.25in}
\caption{The workflow of RTL-OPT for automated benchmarking RTL optimization.
}
\label{benchmark}
% \vspace{-.1in}
\end{figure*}

Based on our aforementioned observations, we propose a new benchmark, \textbf{RTL-OPT}, specifically designed to evaluate LLMs’ ability in RTL design optimization systematically. RTL-OPT consists of 36 handcrafted RTL optimization tasks targeting PPA qualities. 
A key distinguishing feature is that it provides a collection of \textbf{diverse and realistic optimization patterns}, such as \emph{bit-width optimization, precomputation and LUT conversion, operator strength reduction, control simplification, resource sharing, and state encoding optimization}, all derived from proven industry practices.
These patterns capture transformations that truly matter for RTL optimization and remain effective even under advanced synthesis. It sets RTL-OPT apart from prior works that often lacked real optimization impact.
As illustrated in Figure~\ref{benchmark}, each RTL-OPT task provides a pair of RTL codes: a deliberately designed suboptimal (to-be-optimized) version and an optimized version serving as the golden reference. LLMs take the suboptimal code as input and attempts to generate a more optimized RTL code while preserving design functionality. 
Specifically, RTL-OPT provides:   
\ding{182} a set of 36 handcrafted tasks, ensuring comprehensive and representative coverage of real-world design challenges;  
\ding{183} an integrated evaluation framework (Figure~\ref{benchmark}), which automatically verifies functional correctness and compares the ultimate PPA of LLM-optimized designs against the designer-optimized golden reference.

%% file: _txt/2-discussion.tex
% \vspace{-0.2in}
\section{Rethinking the RTL Code Optimization}
\label{sec:rethink}

In this section, we present a comprehensive study examining both existing benchmarks~\cite{yao2024rtlrewriter} for RTL code optimization and our proposed RTL-OPT benchmark across multiple synthesis configurations. This study reveals the inherent limitations of relying solely on overly contrived design examples in conventional benchmarks.

\subsection{Impact of Synthesis Process on RTL Evaluation}
\label{sec:discussion:synthesis}

According to our study, we point out that the evaluation of RTL optimization (i.e., judging which RTL code leads to better PPA) is not a straightforward task. \textbf{One primary reason is that the ultimate design quality also depends on the synthesis process}.
% , which converts the RTL code to the circuit implementation. The synthesis process not only affects the ultimate PPA values, but also the comparison result between a pair of RTL codes.   
Differences in \emph{synthesis tools}, \emph{optimization modes}, and \emph{timing constraints} can all significantly affect whether and how structural differences in RTL code are reflected in the final implementation.

\textbf{Effect of Synthesis Tool.}
% \ly{Synthesis tools can be broadly categorized into commercial and open-source options, with Synopsys Design Compiler (DC)~\cite{designcompiler} and Yosys~\cite{wolf2013yosys} being the most widely used representatives in each category.  
% DC is an industry-standard tool offering advanced optimization capabilities and robust handling of complex RTL constructs. In contrast, Yosys is an open-source weaker alternative valued in academic research. 
% These tools implement varying optimization strategies and heuristics, so the same RTL may produce significantly different outcomes depending on the chosen tool, directly influencing how design quality is perceived.}
Different tools adopt distinct optimization strategies and heuristics. Commercial tools such as Synopsys Design Compiler (DC)~\cite{designcompiler} typically perform more aggressive and sophisticated optimizations, while open-source tools like Yosys~\cite{wolf2013yosys} provide a weaker alternative valued in academic research. As a result, the same pair of RTL codes may exhibit different levels of quality differentiation depending on the synthesis tool used.

\textbf{Effect of Compile Mode.}
% Commercial tools like Synopsys DC support multiple compile modes. For instance, \texttt{compile\_ultra} applies more aggressive and advanced logic optimizations compared to the basic \texttt{compile} mode. \ly{Its aggressive optimization by flattening or restructuring logic tends to obscure fine-grained RTL differences.}
Commercial tools support multiple compilation modes with varying optimization aggressiveness. More advanced modes (e.g., \texttt{compile\_ultra} in DC) tend to aggressively restructure logic, which can obscure fine-grained RTL differences.

\textbf{Effect of Clock Period Constraints.}
The target clock period also shapes synthesis behavior. Tight constraints often lead to aggressive timing-driven optimizations, while relaxed constraints may reduce differentiation between RTL variants. Choosing a realistic and consistent timing target is important for fair and interpretable evaluation of RTL code.

\subsection{Inspection of Existing Benchmark}
\label{sec:discussion:rtlrewriter}

\ly{The existing benchmark~\cite{yao2024rtlrewriter} provides multiple pairs of \emph{suboptimal} and \emph{human-optimized} RTL designs, along with additional RTL code generated by their LLM-based optimization experiments.} Surprisingly, our study reveals that: 1) Both the \emph{human-optimized} RTL designs and the \emph{LLM-optimized} RTL designs from \cite{yao2024rtlrewriter} often fail to outperform their corresponding \emph{suboptimal} counterparts after synthesis. In many cases, they are essentially the same or even worse, particularly when advanced synthesis options are applied. 2) We observe clearly different impacts on ultimate PPAs between different synthesis tools: commercial tool DC with strong optimization capabilities tend to eliminate the differences between suboptimal and optimized RTL, while open-source Yosys often exaggerates them. Together, these results suggest that the existing benchmark does not reliably reflect true improvements in RTL code.

% Surprisingly, our experiments show that both the \emph{human-optimized} RTL designs and the \emph{LLM-optimized} RTL designs from RTLRewriter are often the same as, or even worse than, their corresponding \emph{suboptimal} counterparts after synthesis—especially when advanced synthesis options are used. This trend holds across different subsets of the benchmark and for both commercial and open-source synthesis tools.
\ly{The evaluation results of existing benchmarks are shown in Table~\ref{tab:measure1} and~\ref{tab:measure2}.} 
% To better understand the evaluation of RTL optimization, Table~\ref{tab:measure1} and Table~\ref{tab:measure2} present our comprehensive experiment of the existing benchmark~\cite{yao2024rtlrewriter} under different synthesis settings. 
We carefully inspect and evaluate all 43 pairs of RTL code\footnote{\ly{As shown in Table~\ref{tab:measure1}, paper of~\cite{yao2024rtlrewriter} and SymRTLO~\cite{wang2025symrtlo} are different subsets of the Benchmark~\cite{yao2024rtlrewriter}. We only successfully synthesized 43 cases out of the 54 pairs of RTL code from the benchmark~\cite{yao2024rtlrewriter}. For the others, we synthesized 12 out of 14 pairs and 13 out of 16 pairs, respectively. These synthesis failures in the original benchmarks are mainly caused by Verilog syntax errors.}}
% \footnote{As shown in Table~\ref{tab:measure1}, we only successfully synthesized 43 cases out of the 54 pairs of RTL code from the benchmark~\cite{yao2024rtlrewriter}. For the others, we synthesized 12 out of 14 pairs and 13 out of 16 pairs, respectively. These synthesis failures are mainly caused by Verilog syntax errors, such as assigning to \texttt{wire} variables inside \texttt{always} blocks, which are flagged by DC during parsing. and the subset of 12 and 13 pairs selected by the papers of~\cite{yao2024rtlrewriter} and \cite{wang2025symrtlo}, respectively.}
from the whole benchmark released in~\cite{yao2024rtlrewriter}. Specifically, Table~\ref{tab:measure1} compares each pair of \emph{suboptimal} and \emph{human-optimized} designs, both from the original benchmark. %It evaluates all 54 cases of the original benchmark. 
% These evaluations are carefully conducted with different synthesis processes, including synthesis tools (i.e., DC vs. Yosys), synthesis mode (compile vs. compile\_ultra), and synthesis setups (i.e., clock period). 
% The NanGate45~\cite{NanGate} library is adopted for the technology mapping of all synthesis processes. 
We evaluate whether the \emph{human-optimized} reference is actually better, worse, or the same compared with its \emph{suboptimal} RTL counterpart after synthesis\footnote{The details of the synthesis process, tools, and PPA metrics used for evaluating are provided in Section~\ref{sec:evaluation_methodology}.}. 
% For DC, we evaluate PPA in the reported power, area, total negative slack (TNS), and worst negative slack (WNS). 
In addition, there may exist a ``trade-off'' result in the PPA comparison, indicating improvement in one PPA metric while degradation in the other. As for Yosys, similar to prior works, we only compare the number of cells. In Table~\ref{tab:measure1}, only 13 \emph{human-optimized} RTL~\cite{yao2024rtlrewriter} out of 43 cases are better than their \emph{suboptimal} counterparts with \texttt{compile\_ultra}. This number rises to 16 with \texttt{compile} and further to 24 with Yosys. 
% A similar trend is observed in the number of pairs resulting in the same PPA: 21 out of 43 pairs have the same PPA with \texttt{compile\_ultra}, dropping to 13 with Yosys. 
Many \emph{human-optimized} RTLs are no better than \emph{suboptimal} RTL, particularly with advanced synthesis options. \ly{It validates that commercial tools can eliminate many contrived inefficiencies, while open-source tools often retain them, highlighting a clear discrepancy.}

\begin{table*}[t]
\vspace{-.3in}
\centering
\resizebox{0.9\textwidth}{!}{
    \renewcommand{\arraystretch}{1}
    \setlength{\tabcolsep}{4pt}
\begin{tabular}{c|c|ccc|cccc|cccc}\toprule
\multirow{2}{*}{\textbf{Benchmark}} &
\multirow{2}{*}{\textbf{Total}} &
  \multicolumn{3}{c|}{\textbf{Yosys}} &
  \multicolumn{4}{c|}{\textbf{DC (compile, clk = 1ns | \gray{0.1ns})}} &
  \multicolumn{4}{c}{\textbf{DC (compile\_ultra, clk = 1ns | \gray{0.1ns})}} \\
 & 
 &
  \textit{\textbf{same}} &
  \textit{\textbf{worse}} &
  \textit{\textbf{\yao{better}}} &
  \textit{\textbf{same}} &
  \textit{\textbf{trade-off}} &
  \textit{\textbf{worse}} &
  \textit{\textbf{\yao{better}}} &
  \textit{\textbf{same}} &
  \textit{\textbf{trade-off}} &
  \textit{\textbf{worse}} &
  \textit{\textbf{\yao{better}}} \\ \midrule\midrule
Benchmark of~\cite{yao2024rtlrewriter}    & 43 & 13 & 6 & 24 & 13 | \gray{13} & 7 | \gray{9} & 7 | \gray{6} & 16 | \gray{15} &  21 | \gray{22} & 1 | \gray{3} & 8 | \gray{7} & 13 | \gray{11} \\
Paper of~\cite{yao2024rtlrewriter}    & 12 & 1 & 0 & 11 & 1 | \gray{1} & 4 | \gray{4}& 2 | \gray{1}& 5 | \gray{6} & 4 | \gray{4}& 1 | \gray{1}& 3 | \gray{1}& 4 | \gray{6}\\
SymRTLO~\cite{wang2025symrtlo}     & 13 & 2 & 1 & 10 & 2 | \gray{2} & 2 | \gray{2} & 1 | \gray{2} & 8 | \gray{7} & 4 | \gray{4} & 1 | \gray{0} & 3 | \gray{3} & 5 | \gray{6} \\
\hline
\textbf{RTL-OPT}   & 36 & 3 & 0 & 33 & 0 | \gray{0} & 6 | \gray{13} & 0 | \gray{0} & 30 | \gray{23} & 0 | \gray{0} & 1 | \gray{12} & 0 | \gray{0} & 35 | \gray{24} \\
\bottomrule
\end{tabular}
}
\caption{Comparison between each pair of \emph{suboptimal} and \emph{human-optimized} designs from~\cite{yao2024rtlrewriter} and RTL-OPT (this work).
RTL-OPT shows consistent improvements (35 out of 36 better under \texttt{compile\_ultra} clk = 1ns), matching the expectation that expert-optimized RTL should outperform suboptimal versions. 
In contrast, prior benchmarks often show little or no improvement under stronger synthesis settings, indicating limited reliability for benchmarking RTL optimization.
}
\label{tab:measure1}
\vspace{-0.1 in}
\end{table*}

\begin{table*}[htpb]
\resizebox{0.9\textwidth}{!}{
    \renewcommand{\arraystretch}{1}
    \setlength{\tabcolsep}{4pt}
\begin{tabular}{c|c|ccc|cccc|cccc}\toprule

\multirow{2}{*}{\textbf{LLM Solution}} &
\multirow{2}{*}{\textbf{Total}} &
  \multicolumn{3}{c|}{\textbf{Yosys}} &
  \multicolumn{4}{c|}{\textbf{DC (compile, clk = 1ns)}} &
  \multicolumn{4}{c}{\textbf{DC (compile\_ultra, clk = 1ns)}} \\
 & 
 &
  \textit{\textbf{same}} &
  \textit{\textbf{worse}} &
  \textit{\textbf{\yao{better}}} &
  \textit{\textbf{same}} &
  \textit{\textbf{trade-off}} &
  \textit{\textbf{worse}} &
  \textit{\textbf{\yao{better}}} &
  \textit{\textbf{same}} &
  \textit{\textbf{trade-off}} &
  \textit{\textbf{worse}} &
  \textit{\textbf{\yao{better}}} \\ \midrule\midrule

% \textbf{GPT-4.0}      & 13\,/\,14 & 4 & 1 & 8 & 3 & 3 & 2 & 5 & 4 & 4 & 2 & 3          \\
GPT-4.0        & 13 & 4 & 3 & 6 & 3 & 3 & 2 & 5 & 4 & 4 & 2 & 3          \\
% \textbf{Claude3.0}   & 9/9            &     2        & 2 &           2         &          3          \\
% \textbf{RTLCoder}    & 2/2            &      0      & 0  &            1        &          1          \\
% \textbf{VeriGen}     & 1            &        0     & 1 &          0          &             0       \\
Model~\cite{yao2024rtlrewriter} & 12 & 3 & 1 & 8 & 3 & 4 & 0 & 5 & 3 & 6 & 0 & 3         \\ \bottomrule
\end{tabular}
}
\caption{Comparison between \emph{suboptimal} and \emph{LLM-optimized} designs released from~\cite{yao2024rtlrewriter}. 
% Only 14 LLM-optimized designs used in the paper~\cite{yao2024rtlrewriter} are released. 
Similar to Table~\ref{tab:measure1}, we evaluate whether the LLM-optimized reference is \emph{better}, \emph{worse}, or \emph{same} compared with its suboptimal RTL under different synthesis tools.
}
\label{tab:measure2}
\vspace{-.2in}
\end{table*}

% To further explore how synthesis configuration impacts evaluation, 
We also extend our evaluation under different clock constraints by setting a tighter timing target of clock period = 0.1ns. The corresponding results are shown as the \gray{gray} entries in Table~\ref{tab:measure1}.
% \footnote{The timing constraint also applies to \emph{in2reg}, \emph{reg2out}, and \emph{in2out} paths.}. 
When using the same synthesis modes (i.e., \texttt{compile} or \texttt{compile\_ultra}), using a tighter timing constraint leads to slightly more cases with PPA \emph{trade-offs} and even less actually \emph{better} RTL code. 

Table~\ref{tab:measure2} further compares \emph{LLM-optimized} designs with suboptimal designs directly released by prior work~\cite{yao2024rtlrewriter}. In Table~\ref{tab:measure2}, for both GPT-4.0 and model proposed by~\cite{yao2024rtlrewriter},
% \footnote{We collect the GPT-4 and RTLRewriter-optimized RTL codes from the GitHub repo in~\cite{yao2024rtlrewriter}, according to the file name of `XXX\_gpt4.v' and `XXX\_ours.v'. `XXX' stands for the design name.}
only 3 \emph{LLM-optimized} RTL out of 12 cases are actually better than their \emph{suboptimal} counterparts with \texttt{compile\_ultra}. The number rises to 5 out of 12 with \texttt{compile}. 
% It further rises to 8 out of 12 when Yosys is applied. 
In summary, many \emph{LLM-optimized} RTLs turn out to be no better than \emph{suboptimal} RTL, especially with advanced synthesis options. 
% In addition, the overall performance of GPT-4.0 and the model proposed by~\cite{yao2024rtlrewriter} in RTL code optimization is similar, according to our evaluation\footnote{Since prior work~\cite{yao2024rtlrewriter} does not release their specific Yosys flow or describe their technology library, their synthesis process may differ from ours, possibly leading to sligntly different evaluation results.}. 
\ly{It indicates that under strict synthesis flows, the benchmark shows limited effectiveness.}

\subsection{Same Inspection of Our Benchmark}

In Table~\ref{tab:measure1}, we also evaluate our proposed benchmark, RTL-OPT, using the same setup. 
% Details of the benchmark are covered in Section~\ref{sec:dataset}. 
According to Table~\ref{tab:measure1}, 35 out of 36 \emph{human-optimized} RTL codes in RTL-OPT are better when \texttt{compile\_ultra} is adopted, and 33 out of 36 for Yosys. In Table~\ref{tab:measure1}, with a tight timing constraint, 23 cases remain better while 13 result in PPA trade-offs, with no cases achieving the same ultimate PPA. Compared to the previous benchmark~\cite{yao2024rtlrewriter}, RTL-OPT shows significant improvements in evaluating RTL designs. 
% Specifically, under Yosys evaluation, 13 cases in the previous benchmark are equivalent and 6 are worse. Similarly, under the DC compile ultra setting, only 13 cases outperform the suboptimal RTL codes, with 21 showing no difference and 8 being worse. 
% This indicates a notable enhancement in the quality of human-optimized RTL codes.

\ly{This clear validation of RTL-OPT’s benchmark quality arises from its design philosophy: it provides genuinely \emph{suboptimal} RTL implementations with meaningful room for improvement, rather than contrived inefficiencies that synthesis tools can remove.}

%% file: _txt/3-dataset.tex
% \vspace{-0.1 in}

\section{RTL-OPT Benchmark}
\label{sec:dataset}

% Describe the dataset: basic setup (a pair of suboptimal and optimal RTL code), carefully handcrafted code by engineers, number of designs, type of designs, (statistics: range of size, area, ...), \textbf{[types of difference between suboptimal and optimized RTL code]}, \textbf{some examples}. Quote Table 2, introduce the gap. (We shared all files: RTL code, netlist, measured PPA of suboptimal, optimal, LLM-generated solution). 

% Describe the evaluation: introduce how synthesis works, introduce PPA metrics, and synthesis tools (DC is a much better optimization tool over Yosys), mention the trade-offs among PPAs. Technology node (library), synthesis tools, simulation tools (PrimeTime, PrimePower), functionality verification (Formality/VCS). 

% We describe the dataset construction process, highlight the diverse optimization patterns demonstrated in the examples, and present benchmark results under different synthesis tools. We also detail the evaluation platform used to assess the quality of the design in terms of PPA.

In this section, we present RTL-OPT, a benchmark designed for evaluating RTL code optimization with LLMs. RTL-OPT provides realistic suboptimal and optimized RTL pairs handcrafted by experts, ensuring genuine inefficiencies and meaningful golden references. Covering diverse design types and evaluated with both commercial and open-source tools, it offers a robust and practical resource for advancing RTL optimization research.

% \vspace{-0.1 in}
\subsection{Benchmark Description}

The RTL-OPT consists of 36 RTL design optimization tasks. Each task provides a pair of RTL codes: a suboptimal version and a corresponding designer-optimized version, implementing the same functionality. All designs are manually written by hardware engineers to reflect realistic coding styles and optimization practices, with the optimized RTL serving as the golden reference for human-optimized PPA quality. The suboptimal RTL is not arbitrarily degraded; it represents a valid, functionally correct design that omits specific optimization opportunities. This setup creates meaningful optimization gaps and practical scenarios encountered in the semiconductor industry.

%  providing better hardware

The 36 provided design tasks cover a variety of design types, including arithmetic units, control logic, finite state machines (FSMs), and pipelined datapaths. These designs vary in size and complexity, with logic area ranging from \textbf{14} to \textbf{20K} cells and synthesized area ranging from \textbf{15} to \textbf{19K}~$\mu m^2$. This diversity ensures that the benchmark is representative of practical RTL design tasks.

Table~\ref{tab:general} summarizes the average PPA metrics of RTL-OPT provided suboptimal and optimized designs across 36 benchmarks. The results include synthesis and evaluation using both commercial DC and open-source Yosys tools. From the table, it is evident that the suboptimal designs exhibit significantly higher PPA values compared to the optimized designs. 
These results indicate that the suboptimal designs have substantial room for improvement even under DC, highlighting the effectiveness of RTL-OPT as a benchmark for optimizing the PPA of RTL code in LLMs.

RTL-OPT is fully open-sourced and provides the following artifacts to support benchmarking the RTL optimization capabilities of LLMs: (1) 36 carefully designed RTL code pairs; (2) Corresponding synthesized netlists from commercial synthesis tools; (3) Detailed PPA reports from electronic design automation (EDA) tools for both suboptimal and optimized designs; (4) A complete toolchain flow, including standard-cell library, scripts for synthesis, simulation, and functional verification, which can also verify the correctness of the rewritten code by LLMs.

% Table~\ref{tab:result_1} shows the evaluated PPA of all 36 pairs of suboptimal and optimized designs from RTL-OPT, using both \textbf{DC} and \textbf{Yosys}. Due to its size, this table is now in Appendix~\ref{sec:ppa-1ns}. Detailed explanation of our evaluation methodology, synthesis process, and the PPA metrics are provided in Section~\ref{sec:evaluation_methodology}. 

%These pairs consist of the suboptimal designs and the optimized designs, evaluated using two synthesis tools: commercial \textbf{DC} and open-source \textbf{Yosys}. 

%The results from DC include key metrics such as cell count, area, power, worst negative slack (WNS), and total negative slack (TNS). Yosys provides area-related metrics consistent with those used in RTLRewriter, such as the number of wires and cells. In both tools, the optimized versions consistently outperform the suboptimal versions in terms of PPA. This evaluation with both tools ensures a more robust and realistic benchmark, highlighting the superior performance of the RTL-OPT dataset in evaluating the quality of LLM-optimized RTL designs.

%\subsection{Dataset Details: PPA Qualities of Provided suboptimal and Opti in 
  %compare the optimization performance across several design pairmal Designs}
%The benchmark results, showns. 

% \iffalse
\begin{table}[!t]
\resizebox{0.49\textwidth}{!}{
    \renewcommand{\arraystretch}{1.2}
\begin{tabular}{c|ccccc|ccc}
\toprule
\multicolumn{9}{c}{\textbf{RTL-OPT Suboptimal Designs}} \\ \midrule
\multirow{2}{*}{\textbf{Metrics}} & \multicolumn{5}{c|}{\textbf{DC Results}} & \multicolumn{3}{c}{\textbf{Yosys Result}} \\ \cline{2-9}
 &  \textit{\textbf{Cells}} &
  \textit{\textbf{Area}} &
 \textit{\textbf{\makecell{Power \\ (mW)}}} &
\textit{\textbf{\makecell{WNS \\ (ns)}}} &
\textit{\textbf{\makecell{TNS \\ (ns)}}} & \textit{\textbf{Wires}} & \textit{\textbf{Cells}} & \textit{\textbf{Area}} \\ \midrule
% \textbf{Medium} &  &  &  &  &  &  &  &  \\
% \textbf{Max} &  &  &  &  &  &  &  &  \\
\textbf{Average}  & 1226.8 & 1337.4 & 14.0 & 0.0 & 0.0 & 636.9 & 360.4 & 559.7  \\ 
\bottomrule
\toprule
\multicolumn{9}{c}{\textbf{RTL-OPT Optimized Designs}} \\ \midrule
\multirow{2}{*}{\textbf{Metrics}} & \multicolumn{5}{c|}{\textbf{DC Results}} & \multicolumn{3}{c}{\textbf{Yosys Result}} \\ \cline{2-9}
 &  \textit{\textbf{Cells}} &
  \textit{\textbf{Area}} &
 \textit{\textbf{\makecell{Power \\ (mW)}}} &
\textit{\textbf{\makecell{WNS \\ (ns)}}} &
\textit{\textbf{\makecell{TNS \\ (ns)}}} & \textit{\textbf{Wires}} & \textit{\textbf{Cells}} & \textit{\textbf{Area}} \\ \midrule
% \textbf{Medium} &  &  &  &  &  &  &  &  \\
% \textbf{Max} &  &  &  &  &  &  &  &  \\
\textbf{Average} & 901.8 & 1047.7 & 9.1 & 0.0 & 0.0 & 597.4 & 313.7 & 503.1\\
\midrule\midrule
\textbf{\makecell{Improvement \\ (\%)}} & \textbf{36.04} & \textbf{27.64} & \textbf{53.37} & \textbf{-} & \textbf{-} & \textbf{6.62} & \textbf{14.88} & \textbf{11.25} \\
\bottomrule
\end{tabular}
}
\caption{Average PPA comparison of RTL-OPT provided suboptimal vs. optimized designs. Both commercial DC and open-source Yosys were used for RTL synthesis and PPA evaluation. 
% Trade-offs exist across PPA metrics, and smaller values indicate better performance. 
We only show the average values across all designs due to the space limit, and detailed results are included in Github. }
\label{tab:general}
\vspace{-.15in}
\end{table}

% \fi

\subsection{RTL-OPT Analysis: Optimization Patterns}
\label{sec:intro_b}

The \emph{optimization patterns}, which provide optimization opportunities, are derived from proven industry-standard RTL coding practices that have a direct impact on the quality of logic synthesis. These patterns represent how specific RTL-level modifications ultimately affect downstream synthesis outcomes. %spanning the translation, optimization, and technology mapping phases. 
%Rather than arbitrary or synthetic changes, each optimization pattern is grounded in real-world strategies that hardware engineers commonly apply to achieve PPA objectives. %We classify these optimization patterns into several distinct categories, each representing a specific design strategy aimed at improving synthesis efficiency or resource utilization. Multiple RTL examples within the RTL-OPT dataset illustrate the application of these strategies. 
The key optimization pattern types in the RTL-OPT benchmark are summarized as follows:

\begin{itemize}
    \item \textbf{Bit-width Optimization:} Reducing register and wire widths where full precision is not necessary, optimizing both area and power consumption.
    \item \textbf{Precomputation \& LUT Conversion:} Replacing runtime arithmetic operations with precomputed lookup tables to eliminate complex logic units.
    \item \textbf{Operator Strength Reduction:} Substituting high-cost operators with simpler equivalents through bit manipulation.
    \item \textbf{Control Simplification:} Flattening nested finite state machines (FSMs) or reducing unnecessary states, streamlining control logic, and improving both area and timing.
    \item \textbf{Resource Sharing:} Consolidating duplicate logic across different cycles to maximize hardware resource efficiency.
    \item \textbf{State Encoding Optimization:} Selecting optimal state encoding schemes (One-hot, Gray, Binary) based on state count to balance power, area, and timing.
    
\end{itemize}

By integrating optimization patterns across a diverse range of RTL designs, RTL-OPT generates its realistic yet challenging benchmark for LLM-assisted RTL code optimization: enhancing PPA metrics of optimized code while maintaining functional correctness.   

%for evaluating the ability of LLMs to identify and implement effective improvements to suboptimal RTL code—enhancing PPA metrics while maintaining functional correctness.

% \begin{figure*}[!t]
% \centering
% \includegraphics[width=0.88\textwidth]{_fig/example.png}
% \vspace{-.2in}
% \caption{Comparison of suboptimal and optimized RTL code examples in RTL-OPT. 
% % The suboptimal code we provide is realistic and reasonable. The optimized code indeed has better quality when evaluated with the commercial tool DC, reflecting an actual optimization in industrial applications.
% }
% \label{example}
% \vspace{-.1in}
% \end{figure*}

To illustrate these optimization patterns, we provide \emph{one simple code example} from the RTL-OPT in Figure~\ref{fig:intro_b}. This example compares suboptimal and optimized RTL implementations within a specific pattern category, accompanied by discussions on the structural changes and the quantitative improvements observed in downstream PPA metrics.

\textbf{Example:} This example in Figure~\ref{fig:intro_b} demonstrates the optimization pattern of \textbf{precomputation \& LUT conversion}, where real-time arithmetic operations are replaced with precomputed values. In the suboptimal design, the output is dynamically calculated using the \texttt{sel} input, requiring multiplication at each clock cycle. The optimized design replaces this operation with a \texttt{case} statement that directly assigns precomputed values based on the selection. This optimization results in a 14\% reduction in area and a 12\% decrease in power consumption by eliminating arithmetic operations and reducing signal toggling.

% \textbf{Example 2:} This example in Figure~\ref{fig:intro}(b) (right) demonstrates \textbf{bit-width optimization}, where the physical implementation of an arithmetic operation is restructured to minimize resource usage. In the suboptimal design, a 16-bit multiplication is followed by an addition of the least significant bit, resulting in a larger bit-width for intermediate signals. The optimized design reduces the bit-width by truncating the multiplication result to an 8-bit value directly and simplifying the addition operation. This restructuring achieves a 7\% reduction in area and a 6\% decrease in power consumption by reducing the width of intermediate signals and operations.

% \begin{figure}[!t]
% \includegraphics[width=1\textwidth]{_fig/example.pdf}
% \caption{Example.}
% \label{example}
% % \vspace{-.2in}
% \end{figure}

\subsection{Evaluation Methodology and Tools}\label{sec:evaluation_methodology}

%The evaluation of the 

RTL-OPT provides a complete evaluation flow to assess LLMs' optimization capabilities by measuring the PPA of synthesized RTL code. This is achieved through a combination of synthesis, functionality verification, and PPA evaluation, all performed using industry-standard EDA tools. 

% state-of-the-art 
% trade-offs between 

\subsubsection{Synthesis Process}

The logic synthesis process converts the initial RTL code into gate-level netlists, based on which the PPA metrics can be quantitatively evaluated. 
In this work, we mainly employ DC~\cite{designcompiler} for the synthesis of the RTL-OPT benchmark, given its established effectiveness in industrial design flows. 
DC demonstrates superior capabilities in identifying inefficient RTL constructs and optimizing them into more efficient circuit implementations, thereby minimizing sensitivity to the initial code quality.

When benchmark quality is insufficient, DC tends to synthesize both the suboptimal and optimized RTL codes into functionally equivalent gate-level netlists, resulting in identical PPA outcomes. 
This behavior reflects the limited optimization opportunities offered by low-quality benchmarks. 
Conversely, open-source synthesis tools such as Yosys~\cite{wolf2013yosys}, which provide less aggressive optimization, may still produce differing PPA results for such code pairs, potentially overstating the effectiveness of certain code transformations. For completeness, we also provide synthesis results obtained using Yosys to support broader comparative analyses.

The synthesis process also involves the use of a technology library, or cell library, which is a collection of pre-characterized standard cells such as logic gates, flip-flops, and other fundamental components. These cells are designed to meet specific PPA constraints.
While we use Nangate45 \cite{NanGate} in our evaluation, other libraries could also be used, though they typically require a license. The choice of library significantly impacts the RTL optimization process, as it defines the available cells and their performance characteristics, ultimately influencing the design's efficiency.

\subsubsection{Functional Equivalance Verification}
After successful synthesis, the RTL code is ensured to be free of syntax errors, as it can be correctly transformed into a gate-level netlist.
Following synthesis, functional verification is essential to ensure that the optimization steps have not introduced errors. This verification is primarily conducted using \textbf{Synopsys Formality}~\cite{fomal}, which performs functional equivalence checking by rigorously comparing the LLM-optimized RTL against the golden reference to ensure behavioral consistency. However, for optimizations involving timing adjustments, such as pipelining, additional dynamic verification is required. This is performed using \textbf{Synopsys VCS}~\cite{vcs}, which employs comprehensive testbenches to validate the design's behavior under various operating conditions.

This combined approach ensures both logical equivalence and operational reliability of the optimized design. A design is considered functionally valid only if it passes both formal equivalence checking and dynamic verification for timing-critical optimizations.

\subsubsection{PPA Metrics and Trade-offs in Optimization}
To evaluate the quality of the synthesized designs, we analyze them from three aspects: \textbf{Power}, \textbf{Performance}, and \textbf{Area}:

%     \textbf{Power:} The total power consumption of the synthesized design, characterized by the fundamental equation:    $P_{\text{dynamic}} = \alpha C V^2 f$, where $\alpha$ is the switching activity, $C$ is the capacitance, $V$ is the supply voltage, and $f$ is the clock frequency. %This equation captures the dominant factors in the digital circuit's power dissipation.
    
%     \textbf{Performance:} Evaluated through two key timing metrics: \textit{Worst Negative Slack (WNS)}, which represents the largest single timing violation in the design, and \textit{Total Negative Slack (TNS)}, the sum of all timing violations across failing paths. 

%    \textbf{Area}: Characterized by two complementary measures: \textit{Silicon area} (in $\mu m^2$), which indicates the physical implementation footprint, and \textit{Cell count}, the total number of standard cells in the design, providing a basic area estimation that does not account for cell types, placement, or routing overhead.

\textbf{Power:} Total power reported by the synthesis tool, including dynamic and leakage components.

\textbf{Performance:} Timing quality measured by \textit{Worst Negative Slack (WNS)} and \textit{Total Negative Slack (TNS)}.

\textbf{Area:} Physical implementation cost characterized by \textit{silicon area} and \textit{cell count}.

Trade-offs widely exist in these PPA metrics. For instance, optimizing for power may increase area, while minimizing area could compromise power efficiency. A key challenge in RTL optimization is managing these competing goals to achieve an optimal balance based on design constraints.

%% file: _txt/4-experiments.tex
\section{Experiments}\label{sec:exp}

% Describe the performance of different LLMs. 

This section presents the experiments conducted to evaluate the optimization capabilities of different LLMs on the RTL-OPT benchmark. We compare the performance of several LLMs, including \textbf{GPT-4o-mini}, \textbf{Gemini-2.5}, \textbf{Deepseek V3}, and \textbf{Deepseek R1}, in optimizing RTL code. The focus of the experiments is on assessing the optimization in terms of PPA metrics, as well as the functional correctness of the optimized designs.
\ly{The results show that the two Deepseek models demonstrate stronger optimization ability than the other evaluated LLMs.} 
% Detailed tables summarizing PPA performance and functional correctness (Table~\ref{tab:exp}) are included in Appendix~\ref{sec:app:exp}.

% Table~\ref{tab:exp} summarizes the evaluated PPA performance of each LLM-optimized design and compares it with the provided suboptimal RTL and optimized RTL (golden reference). \colorbox[HTML]{B7EFA5}{Green cells} indicate that the PPA is better than the suboptimal RTL, and \colorbox[HTML]{B7EFA5}{\textbf{bold green cells}} indicate that the PPA surpasses the optimized RTL (golden reference). The table also shows the functional correctness after verification (Func column), with \ding{52} and \colorbox[HTML]{FFCCC9}{\ding{56}} representing the verification results. \colorbox[HTML]{FFCCC9}{\ding{56}\ding{56}} indicates that the corresponding design contains syntax errors and fails to pass DC synthesis.

%and the LLM-optimized designs. 
%For each model, the table shows the power consumption, area, and timing results for each design, as well as the functional verification status (Func column). 

% Both table includes the PPA metrics for each design, comparing the suboptimal RTL, optimized RTL (golden reference), and the LLM-optimized designs. For each model, the table shows the power consumption, area, and timing results for each design, as well as the functional verification status (Func column).

% 表格 PPA quality 1ns

\subsection{Summary of Benchmarking Results}
\label{benchmark_result}
% \begin{figure}[!t]
% \includegraphics[width=0.5\textwidth]{_fig/analysis_grouped.png}
% \caption{Comparison of Optimization Performance Across Different LLMs.}
% \label{fig:analysis}
% \vspace{-.1in}
% \end{figure}

%The benchmarking results, as shown in

\begin{figure}[!t]
\centering
\includegraphics[width=0.45\textwidth]{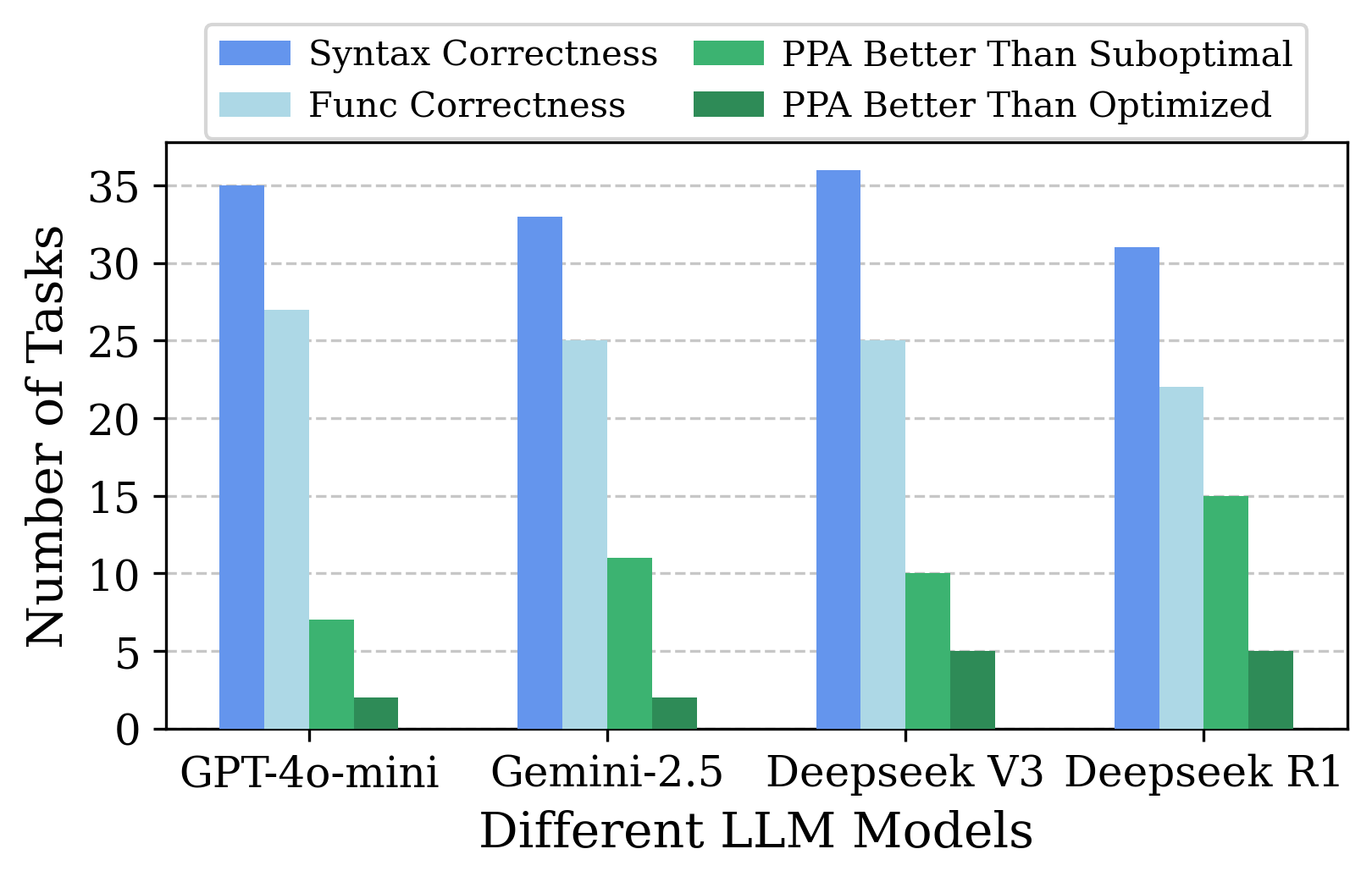}
\vspace{-.05in}
\caption{Comparison of suboptimal and optimized RTL code examples in RTL-OPT. 
% The suboptimal code we provide is realistic and reasonable. The optimized code indeed has better quality when evaluated with the commercial tool DC, reflecting an actual optimization in industrial applications.
}
\label{fig:analysis}
\vspace{-.2in}
\end{figure}

Figure~\ref{fig:analysis} shows a summary of benchmarking results of the four evaluated LLMs. It reveals the syntax correctness, functionality correctness, and post-optimization PPA quality performance of the various LLMs. The overall results highlight that: \ding{182} There is still significant room for LLM to improve in RTL optimization compared to human designers. \ding{183} Our benchmark is designed to be realistic, providing a set of challenging tasks that reflect the complexities encountered in real-world hardware design.

Notably, the overall performance of all LLMs is not very good, reflecting the challenges in our RTL-OPT benchmark. Many LLMs have over 10 optimized cases failed to maintain functionality correctness. Deepseek R1 can successfully optimize about 15 suboptimal designs, and can outperform our human designers' solution for around 5 designs.

When comparing these 4 LLMs, Deepseek R1 generally outperforms the other models in terms of PPA. %achieving substantial improvements in power consumption, area, and timing. 
However, Deepseek R1 also exhibits a higher rate of functional discrepancies compared to the other models. In contrast, models such as GPT-4o-mini and Gemini-2.5, while maintaining high syntax correctness, achieved fewer improvements in PPA. It implies that their optimization strategies are either more conservative or lack effective optimizations.

%This trend underscores the importance of balancing aggressive optimization strategies with the need to maintain functional correctness, particularly in hardware design tasks, where even small changes can lead to significant errors.

\begin{table*}[!t]
\vspace{-.2 in}
    % \centering
\resizebox{1\textwidth}{!}{
\renewcommand{\arraystretch}{0.92}
\begin{tabular}{l|cccccc||cccccc||cccccc}\toprule
\multicolumn{1}{c|}{} &
  \multicolumn{6}{c||}{\textbf{GPT-4o-mini}} &
  \multicolumn{6}{c||}{\textbf{Gemini-2.5}}&
  \multicolumn{6}{c}{\textbf{DeepSeek R1}} \\
\multicolumn{1}{c|}{\multirow{-2}{*}{\textbf{Category}}} &
  \textit{\textbf{Cells}} &
  \textit{\textbf{Area}} &
 \textit{\textbf{\makecell{Power \\ (mW)}}} &
\textit{\textbf{\makecell{WNS \\ (ns)}}} &
\textit{\textbf{\makecell{TNS \\ (ns)}}} &
  \textit{\textbf{Check}} &
  \textit{\textbf{Cells}} &
  \textit{\textbf{Area}} &
  \textit{\textbf{\makecell{Power \\ (mW)}}} &
\textit{\textbf{\makecell{WNS \\ (ns)}}} &
\textit{\textbf{\makecell{TNS \\ (ns)}}} &
  \textit{\textbf{Check}} &
  \textit{\textbf{Cells}} &
  \textit{\textbf{Area}} &
  \textit{\textbf{\makecell{Power \\ (mW)}}} &
\textit{\textbf{\makecell{WNS \\ (ns)}}} &
\textit{\textbf{\makecell{TNS \\ (ns)}}} &
  \textit{\textbf{Check}} \\ \midrule\midrule
\textbf{adder} &
  - &
  - &
  - &
  - &
  - &
  \cellcolor[HTML]{FFCCC9}\ding{56} &
  - &
  - &
  - &
  - &
  - &
  \cellcolor[HTML]{FFCCC9}\ding{56}\ding{56}&
  \cellcolor[HTML]{B7EFA5}433 &
  \cellcolor[HTML]{B7EFA5}511.52 &
  \cellcolor[HTML]{B7EFA5}368.04 &
  \cellcolor[HTML]{B7EFA5}0.00 &
  \cellcolor[HTML]{B7EFA5}0.00 &
  \ding{52} \\
\textbf{adder\_select} &
  450 &
  439.17 &
  264.34 &
  0.00 &
  0.00 &
  \ding{52} &
  450 &
  439.17 &
  264.34 &
  0.00 &
  0.00 &
  \ding{52}&
  \cellcolor[HTML]{B7EFA5}306 &
  \cellcolor[HTML]{B7EFA5}350.06 &
  \cellcolor[HTML]{B7EFA5}225.74 &
  \cellcolor[HTML]{B7EFA5}0.00 &
  \cellcolor[HTML]{B7EFA5}0.00 &
  \ding{52} \\
\textbf{alu\_64bit} &
  1683 &
  1645.21 &
  887.78 &
  0.00 &
  0.00 &
  \ding{52} &
  1683 &
  1645.21 &
  887.78 &
  0.00 &
  0.00 &
  \ding{52} &
  1697 &
  1658.24 &
  896.08 &
  0.00 &
  0.00 &
  \ding{52}\\
\textbf{alu\_8bit} &
  146 &
  179.55 &
  79.58 &
  0.00 &
  0.00 &
  \ding{52} &
  146 &
  179.55 &
  79.58 &
  0.00 &
  0.00 &
  \ding{52}&
  146 &
  179.55 &
  79.58 &
  0.00 &
  0.00 &
  \cellcolor[HTML]{FFCCC9}\ding{56}  \\
\textbf{calculation} &
  - &
  - &
  - &
  - &
  - &
  \cellcolor[HTML]{FFCCC9}\ding{56} &
  888 &
  1044.32 &
  788.67 &
  0.61 &
  5.66 &
  \ding{52} &
  \cellcolor[HTML]{B7EFA5}755 &
  \cellcolor[HTML]{B7EFA5}859.98 &
  \cellcolor[HTML]{B7EFA5}617.26 &
  \cellcolor[HTML]{B7EFA5}0.60 &
  \cellcolor[HTML]{B7EFA5}4.83 &
  \ding{52} \\
\textbf{comparator} &
  \cellcolor[HTML]{B7EFA5}59 &
  \cellcolor[HTML]{B7EFA5}52.67 &
  \cellcolor[HTML]{B7EFA5}26.44 &
  \cellcolor[HTML]{B7EFA5}0.00 &
  \cellcolor[HTML]{B7EFA5}0.00 &
  \ding{52} &
  \cellcolor[HTML]{B7EFA5}54 &
  \cellcolor[HTML]{B7EFA5}47.61 &
  \cellcolor[HTML]{B7EFA5}24.64 &
  \cellcolor[HTML]{B7EFA5}0.00 &
  \cellcolor[HTML]{B7EFA5}0.00 &
  \ding{52}  &
  46 &
  42.29 &
  20.87 &
  0.00 &
  0.00 &
  \ding{52} \\
\textbf{comparator\_16bit} &
  - &
  - &
  - &
  - &
  - &
  \cellcolor[HTML]{FFCCC9}\ding{56} &
  102 &
  88.31 &
  44.57 &
  0.00 &
  0.00 &
  \ding{52} &
  \cellcolor[HTML]{B7EFA5}\textbf{70} &
  \cellcolor[HTML]{B7EFA5}\textbf{65.70} &
  \cellcolor[HTML]{B7EFA5}\textbf{30} &
  \cellcolor[HTML]{B7EFA5}\textbf{0.00} &
  \cellcolor[HTML]{B7EFA5}\textbf{0.00} &
  \ding{52}\\
\textbf{comparator\_2bit} &
  10 &
  9.58 &
  4.51 &
  0.00 &
  0.00 &
  \ding{52} &
  \cellcolor[HTML]{B7EFA5}10 &
  \cellcolor[HTML]{B7EFA5}8.78 &
  \cellcolor[HTML]{B7EFA5}3.82 &
  \cellcolor[HTML]{B7EFA5}0.00 &
  \cellcolor[HTML]{B7EFA5}0.00 &
  \ding{52} &
  \cellcolor[HTML]{B7EFA5}\textbf{9} &
  \cellcolor[HTML]{B7EFA5}\textbf{7.45} &
  \cellcolor[HTML]{B7EFA5}\textbf{3.75} &
  \cellcolor[HTML]{B7EFA5}\textbf{0.00} &
  \cellcolor[HTML]{B7EFA5}\textbf{0.00} &
  \ding{52} \\
\textbf{comparator\_4bit} &
  \cellcolor[HTML]{B7EFA5}21 &
  \cellcolor[HTML]{B7EFA5}18.35 &
  \cellcolor[HTML]{B7EFA5}9.25 &
  \cellcolor[HTML]{B7EFA5}0.00 &
  \cellcolor[HTML]{B7EFA5}0.00 &
  \ding{52} &
  \cellcolor[HTML]{B7EFA5}21 &
  \cellcolor[HTML]{B7EFA5}18.35 &
  \cellcolor[HTML]{B7EFA5}9.25 &
  \cellcolor[HTML]{B7EFA5}0.00 &
  \cellcolor[HTML]{B7EFA5}0.00 &
  \ding{52} &
  - &
  - &
  - &
  - &
  - &
  \cellcolor[HTML]{FFCCC9}\ding{56}\ding{56}\\
\textbf{comparator\_8bit} &
  - &
  - &
  - &
  - &
  - &
  \cellcolor[HTML]{FFCCC9}\ding{56} &
  48 &
  43.36 &
  18.56 &
  0.00 &
  0.00 &
  \ding{52}&
  - &
  - &
  - &
  - &
  - &
  \cellcolor[HTML]{FFCCC9}\ding{56}\ding{56} \\
\textbf{decoder\_6bit} &
  \cellcolor[HTML]{B7EFA5}\textbf{86} &
  \cellcolor[HTML]{B7EFA5}\textbf{71.29} &
  \cellcolor[HTML]{B7EFA5}\textbf{19.76} &
  \cellcolor[HTML]{B7EFA5}\textbf{0.00} &
  \cellcolor[HTML]{B7EFA5}\textbf{0.00} &
  \ding{52} &
  87 &
  71.55 &
  19.54 &
  0.00 &
  0.00 &
  \ding{52}  &
  96 &
  76.61 &
  22.56 &
  0.00 &
  0.00 &
  \cellcolor[HTML]{FFCCC9}\ding{56}\\
\textbf{decoder\_8bit} &
  312 &
  246.85 &
  50.28 &
  0.00 &
  0.00 &
  \ding{52} &
  \cellcolor[HTML]{B7EFA5}\textbf{308} &
  \cellcolor[HTML]{B7EFA5}\textbf{246.85} &
  \cellcolor[HTML]{B7EFA5}\textbf{49.75} &
  \cellcolor[HTML]{B7EFA5}\textbf{0.00} &
  \cellcolor[HTML]{B7EFA5}\textbf{0.00} &
  \ding{52} &
  \cellcolor[HTML]{B7EFA5}\textbf{329} &
  \cellcolor[HTML]{B7EFA5}\textbf{267.33} &
  \cellcolor[HTML]{B7EFA5}\textbf{54.11} &
  \cellcolor[HTML]{B7EFA5}\textbf{0.00} &
  \cellcolor[HTML]{B7EFA5}\textbf{0.00} &
  \ding{52} \\
\textbf{divider\_16bit} &
  3445 &
  3377.67 &
  5030 &
  -3.32 &
  -73.2 &
  \ding{52} &
  \cellcolor[HTML]{B7EFA5}2461 &
  \cellcolor[HTML]{B7EFA5}2372.72 &
  \cellcolor[HTML]{B7EFA5}3750 &
  \cellcolor[HTML]{B7EFA5}7.22 &
  \cellcolor[HTML]{B7EFA5}156. &
  \ding{52} &
  3445 &
  3377.67 &
  5030 &
  3.32 &
  73.2 &
  \ding{52} \\
\textbf{divider\_32bit} &
  14400 &
  14295.90 &
  23800 &
  10.19 &
  468. &
  \ding{52} &
  - &
  - &
  - &
  - &
  - &
  \ding{52} &
  - &
  - &
  - &
  - &
  - &
  \cellcolor[HTML]{FFCCC9}\ding{56}\ding{56}\\
\textbf{divider\_4bit} &
  - &
  - &
  - &
  - &
  - &
  \ding{52} &
  - &
  - &
  - &
  - &
  - &
  \cellcolor[HTML]{FFCCC9}\ding{56} &
  39 &
  40.43 &
  21.58 &
  0.00 &
  0.00 &
  \cellcolor[HTML]{FFCCC9}\ding{56} \\
\textbf{divider\_8bit} &
  571 &
  575.36 &
  681.84 &
  0.50 &
  4.57 &
  \ding{52} &
  \cellcolor[HTML]{B7EFA5}\textbf{266} &
  \cellcolor[HTML]{B7EFA5}\textbf{264.14} &
  \cellcolor[HTML]{B7EFA5}\textbf{188.61} &
  \cellcolor[HTML]{B7EFA5}\textbf{0.51} &
  \cellcolor[HTML]{B7EFA5}\textbf{3.16} &
  \ding{52} &
  - &
  - &
  - &
  - &
  - &
  \cellcolor[HTML]{FFCCC9}\ding{56}\ding{56} \\
\textbf{fsm} &
  89 &
  128.74 &
  92.56 &
  0.00 &
  0.00 &
  \ding{52} &
  - &
  - &
  - &
  - &
  - &
  \cellcolor[HTML]{FFCCC9}\ding{56}\ding{56} &
  \cellcolor[HTML]{B7EFA5}73 &
  \cellcolor[HTML]{B7EFA5}92.04 &
  \cellcolor[HTML]{B7EFA5}51.44 &
  \cellcolor[HTML]{B7EFA5}0.00 &
  \cellcolor[HTML]{B7EFA5}0.00 &
  \ding{52} \\
\textbf{fsm\_encode} &
  287 &
  416.82 &
  337.09 &
  0.00 &
  0.00 &
  \ding{52} &
  \cellcolor[HTML]{B7EFA5}\textbf{151} &
  \cellcolor[HTML]{B7EFA5}\textbf{302.18} &
  \cellcolor[HTML]{B7EFA5}\textbf{273.32} &
  \cellcolor[HTML]{B7EFA5}\textbf{0.00} &
  \cellcolor[HTML]{B7EFA5}\textbf{0.00} &
  \ding{52} &
  \cellcolor[HTML]{B7EFA5}151 &
  \cellcolor[HTML]{B7EFA5}302.18 &
  \cellcolor[HTML]{B7EFA5}273.32 &
  \cellcolor[HTML]{B7EFA5}0.00 &
  \cellcolor[HTML]{B7EFA5}0.00 &
  \ding{52} \\
\textbf{gray} &
  - &
  - &
  - &
  - &
  - &
  \cellcolor[HTML]{FFCCC9}\ding{56} &
  51 &
  69.69 &
  66.78 &
  0.00 &
  0.00 &
  \ding{52} &
  52 &
  70.22 &
  65.86 &
  0.00 &
  0.00 &
  \ding{52} \\
\textbf{mac} &
  - &
  - &
  - &
  - &
  - &
  \cellcolor[HTML]{FFCCC9}\ding{56} &
  - &
  - &
  - &
  - &
  - &
  \cellcolor[HTML]{FFCCC9}\ding{56} &
  352 &
  632.02 &
  787.88 &
  0.00 &
  0.00 &
  \cellcolor[HTML]{FFCCC9}\ding{56} \\
\textbf{mul} &
  315 &
  378.52 &
  421.67 &
  0.05 &
  0.13 &
  \ding{52} &
  315 &
  378.52 &
  421.67 &
  0.05 &
  0.13 &
  \ding{52} &
  315 &
  378.52 &
  421.67 &
  0.05 &
  0.13 &
  \ding{52} \\
\textbf{mul\_sub} &
  \cellcolor[HTML]{B7EFA5}233 &
  \cellcolor[HTML]{B7EFA5}337.02 &
  \cellcolor[HTML]{B7EFA5}256.53 &
  \cellcolor[HTML]{B7EFA5}0.00 &
  \cellcolor[HTML]{B7EFA5}0.00 &
  \ding{52} &
  \cellcolor[HTML]{B7EFA5}234 &
  \cellcolor[HTML]{B7EFA5}341.81 &
  \cellcolor[HTML]{B7EFA5}255.66 &
  \cellcolor[HTML]{B7EFA5}0.00 &
  \cellcolor[HTML]{B7EFA5}0.00 &
  \ding{52} &
  234 &
  336.22 &
  259.8 &
  0.00 &
  0.00 &
  \ding{52} \\
\textbf{mux} &
 - &
  - &
  - &
  - &
  - &
 \cellcolor[HTML]{FFCCC9}\ding{56} &
  - &
  - &
  - &
  - &
  - &
  \cellcolor[HTML]{FFCCC9}\ding{56} &
  - &
  - &
  - &
  - &
  - &
  \cellcolor[HTML]{FFCCC9}\ding{56} \\
\textbf{mux\_encode} &
  - &
  - &
  - &
  - &
  - &
  \cellcolor[HTML]{FFCCC9}\ding{56} &
   - &
  - &
  - &
  - &
  - &
  \cellcolor[HTML]{FFCCC9}\ding{56} &
  \cellcolor[HTML]{B7EFA5}61 &
  \cellcolor[HTML]{B7EFA5}73.42 &
  \cellcolor[HTML]{B7EFA5}35.39 &
  \cellcolor[HTML]{B7EFA5}0.00 &
  \cellcolor[HTML]{B7EFA5}0.00 &
  \ding{52} \\
\textbf{saturating\_add} &
  - &
  - &
  - &
  - &
  - &
  \cellcolor[HTML]{FFCCC9}\ding{56}\ding{56} &
  \cellcolor[HTML]{B7EFA5}18 &
  \cellcolor[HTML]{B7EFA5}67.56 &
  \cellcolor[HTML]{B7EFA5}65.62 &
  \cellcolor[HTML]{B7EFA5}0.00 &
  \cellcolor[HTML]{B7EFA5}0.00 &
  \ding{52} &
  - &
  - &
  - &
  - &
  - &
  \cellcolor[HTML]{FFCCC9}\ding{56}\ding{56} \\
\textbf{selector} &
  \cellcolor[HTML]{B7EFA5}21 &
  \cellcolor[HTML]{B7EFA5}44.69 &
  \cellcolor[HTML]{B7EFA5}39.6 &
  \cellcolor[HTML]{B7EFA5}0.00 &
  \cellcolor[HTML]{B7EFA5}0.00 &
  \ding{52} &
  18 &
  39.37 &
  36.06 &
  0.00 &
  0.00 &
  \ding{52} &
  \cellcolor[HTML]{B7EFA5}18 &
  \cellcolor[HTML]{B7EFA5}39.37 &
  \cellcolor[HTML]{B7EFA5}36.06 &
  \cellcolor[HTML]{B7EFA5}0.00 &
  \cellcolor[HTML]{B7EFA5}0.00 &
  \ding{52} \\
\textbf{sub\_16bit} &
  132 &
  136.19 &
  77.92 &
  0.00 &
  0.00 &
  \ding{52} &
  132 &
  136.19 &
  77.92 &
  0.00 &
  0.00 &
  \ding{52} &
  \cellcolor[HTML]{B7EFA5}\textbf{122} &
  \cellcolor[HTML]{B7EFA5}\textbf{126.88} &
  \cellcolor[HTML]{B7EFA5}\textbf{71.97} &
  \cellcolor[HTML]{B7EFA5}\textbf{0.00} &
  \cellcolor[HTML]{B7EFA5}\textbf{0.00} &
  \ding{52} \\
\textbf{sub\_32bit} &
  \cellcolor[HTML]{B7EFA5}278 &
  \cellcolor[HTML]{B7EFA5}252.17 &
  \cellcolor[HTML]{B7EFA5}146.9 &
  \cellcolor[HTML]{B7EFA5}0.00 &
  \cellcolor[HTML]{B7EFA5}0.00 &
  \ding{52} &
  \cellcolor[HTML]{B7EFA5}278 &
  \cellcolor[HTML]{B7EFA5}252.17 &
  \cellcolor[HTML]{B7EFA5}146.9 &
  \cellcolor[HTML]{B7EFA5}0.00 &
  \cellcolor[HTML]{B7EFA5}0.00 &
  \ding{52} &
  - &
  - &
  - &
  - &
  - &
  \cellcolor[HTML]{FFCCC9}\ding{56} \\
\textbf{sub\_4bit} &
  N/A &
  N/A &
  N/A &
  N/A &
  N/A &
  \ding{52} &
  - &
  - &
  - &
  - &
  - &
  \cellcolor[HTML]{FFCCC9}\ding{56} &
  \cellcolor[HTML]{B7EFA5}14 &
  \cellcolor[HTML]{B7EFA5}19.42 &
  \cellcolor[HTML]{B7EFA5}10.61 &
  \cellcolor[HTML]{B7EFA5}0.00 &
  \cellcolor[HTML]{B7EFA5}0.00 &
  \ding{52} \\
\textbf{sub\_8bit} &
  \cellcolor[HTML]{B7EFA5}27 &
  \cellcolor[HTML]{B7EFA5}41.50 &
  \cellcolor[HTML]{B7EFA5}23.91 &
  \cellcolor[HTML]{B7EFA5}0.00 &
  \cellcolor[HTML]{B7EFA5}0.00 &
  \ding{52} &
  - &
  - &
  - &
  - &
  - &
  \cellcolor[HTML]{FFCCC9}\ding{56} &
  \cellcolor[HTML]{B7EFA5}\textbf{27} &
  \cellcolor[HTML]{B7EFA5}\textbf{41.50} &
  \cellcolor[HTML]{B7EFA5}\textbf{23.91} &
  \cellcolor[HTML]{B7EFA5}\textbf{0.00} &
  \cellcolor[HTML]{B7EFA5}\textbf{0.00} &
  \ding{52} \\
\textbf{add\_sub} &
  124 &
  130.34 &
  101.74 &
  0.00 &
  0.00 &
  \ding{52} &
  124 &
  130.34 &
  101.74 &
  0.00 &
  0.00 &
  \ding{52} &
  - &
  - &
  - &
  - &
  - &
  \cellcolor[HTML]{FFCCC9}\ding{56} \\
\textbf{addr\_calcu} &
  78 &
  131.40 &
  96.07 &
  0.03 &
  0.06 &
  \ding{52} &
  - &
  - &
  - &
  - &
  - &
  \cellcolor[HTML]{FFCCC9}\ding{56} &
  - &
  - &
  - &
  - &
  - &
  \cellcolor[HTML]{FFCCC9}\ding{56} \\
\textbf{mult\_if} &
  10 &
  10.91 &
  3.53 &
  0.00 &
  0.00 &
  \ding{52} &
  10 &
  10.91 &
  3.53 &
  0.00 &
  0.00 &
  \ding{52} &
  - &
  - &
  - &
  - &
  - &
  \cellcolor[HTML]{FFCCC9}\ding{56} \\
\textbf{mux\_large} &
  81 &
  96.82 &
  40.84 &
  0.00 &
  0.00 &
  \ding{52} &
  - &
  - &
  - &
  - &
  - &
  \cellcolor[HTML]{FFCCC9}\ding{56} &
  \cellcolor[HTML]{B7EFA5}89 &
  \cellcolor[HTML]{B7EFA5}100.02 &
  \cellcolor[HTML]{B7EFA5}50.57 &
  \cellcolor[HTML]{B7EFA5}0.00 &
  \cellcolor[HTML]{B7EFA5}0.00 &
  \ding{52} \\
\textbf{register} &
  \cellcolor[HTML]{B7EFA5}4625 &
  \cellcolor[HTML]{B7EFA5}9226.74 &
  \cellcolor[HTML]{B7EFA5}7540 &
  \cellcolor[HTML]{B7EFA5}0.00 &
  \cellcolor[HTML]{B7EFA5}0.00 &
  \ding{52} &
  - &
  - &
  - &
  - &
  - &
  \cellcolor[HTML]{FFCCC9}\ding{56}\ding{56} &
  3731 &
  8745.55 &
  7710 &
  0.00 &
  0.00 &
  \ding{52} \\
\textbf{ticket\_machine} &
  - &
  - &
  - &
  - &
  - &
  \cellcolor[HTML]{FFCCC9}\ding{56} &
  - &
  - &
  - &
  - &
  - &
  \cellcolor[HTML]{FFCCC9}\ding{56} &
  - &
  - &
  - &
  - &
  - &
  \cellcolor[HTML]{FFCCC9}\ding{56}\\ \bottomrule
  \end{tabular}
}

% \vspace{-.0in}
\caption{PPA quality (DC \textbf{\textit{compile\_ultra, 1ns}}) and functional correctness for all designs optimized by GPT-4o-mini, Gemini-2.5, and Deepseek R1, using the RTL-OPT benchmark.}
\vspace{-.2in}
\label{tab:exp}
\end{table*}

\subsection{Detailed Benchmarking Results}
\label{sec:4.2}
%\subsubsection{Trade-off Between Optimization and Functionality}

One observation from the evaluation results in Figure~\ref{fig:analysis} is the trade-off between optimization and functionality. While Deepseek R1 showed the most significant improvements in PPA, it was also the most prone to introducing functional errors. In contrast, GPT-4o-mini and Gemini-2.5 exhibited a more balanced approach, optimizing PPA while maintaining syntax correctness. %across all design tasks.
This indicates that Deepseek R1's aggressive optimization, though effective, tends to increase error, especially in designs with complex timing or control logic.
% For instance, in \texttt{sub\_32bit}, Deepseek R1 achieved a 9\% reduction in area and a 13\% reduction in power, but it introduced functional discrepancies. 
Conversely, GPT-4o-mini and Gemini-2.5, while less aggressive, maintained functional correctness and achieved meaningful PPA improvements. 
% We provide more detailed LLM evaluation results in Appendix~\ref{appendix:llm-eval}.

For example, in the \texttt{sub\_32bit}, Deepseek R1 achieved a 9\% reduction in area and a 13\% reduction in power, but it also resulted in functional discrepancies. This suggests that Deepseek R1's aggressive optimization approach, while effective in improving PPA, may also increase the likelihood of errors, particularly in designs with complex timing or control logic.

Table~\ref{tab:exp} summarize the evaluated PPA performance of each LLM-optimized design and compare it with the provided suboptimal RTL and optimized RTL (golden reference). \colorbox[HTML]{B7EFA5}{Green cells} indicate that the PPA is better than the suboptimal RTL, and \colorbox[HTML]{B7EFA5}{\textbf{bold green cells}} indicate that the PPA surpasses the optimized RTL. The table also shows the functional correctness after verification (Func column), with \ding{52} and \colorbox[HTML]{FFCCC9}{\ding{56}} representing the verification results. \colorbox[HTML]{FFCCC9}{\ding{56}\ding{56}} indicates that the corresponding design contains syntax errors and fails to pass DC synthesis.

For the experimental results, we summarize the performance of four LLMs across syntax correctness, functional correctness, and PPA improvement. 
1) \textbf{GPT-4o-mini} achieves good correctness, with a syntax correctness rate of 97.2\% and functional correctness of 75\%. Though only 19.4\% of its generated code achieving better PPA than the suboptimal version. 2) Similarly, \textbf{Gemini-2.5} exhibited the same trend as GPT-4o-mini: relatively high functional correctness but low performance in PPA optimization. 3) For \textbf{Deepseek V3}, it gets the highest syntax correctness of 100\%, and the same functional correctness of 69.4\% with Gemini-2.5. It achieved a balanced performance across all metrics. 4) In contrast, \textbf{Deepseek R1}, with a syntax correctness rate of 86.1\% and functional correctness of 61.1\%, produced 41.7\% of the code with better PPA than the suboptimal version, and 13.9\% better than designer solutions, despite its lower functional correctness.

% Overall, these results highlight the importance of balancing aggressive optimization with functqional verification, particularly in hardware design tasks where even small changes can lead to significant errors.

Beyond quantitative results, we also randomly inspected 40 cases where LLM-optimized designs passed syntax checks but failed functional verification. We observed three main failure modes: \textbf{control logic inconsistencies} (e.g., incorrect Boolean conditions in comparators), \textbf{overly aggressive pipelining} (e.g., violating latency requirements in FSMs), and \textbf{improper resource sharing} (e.g., stale data due to register reuse). These results highlight that LLM errors often stem from subtle design semantics rather than surface-level syntax issues. 
% We provide the details in Appendix~\ref{sec:app:quan}.

%% file: _txt/5-conclusion.tex
% \section{Discussion: Beyond Local RTL Optimization}

% Our findings suggest that current LLMs, when tasked with optimizing RTL code, tend to make localized edits based on the given suboptimal design. This behavior reflects a form of \emph{local optimization}, where the model focuses on refining or simplifying the existing structure rather than exploring fundamentally different design alternatives.

% While such incremental improvements are valuable and often sufficient, they may fall short in cases where deeper architectural changes are required to achieve significant PPA gains. For instance, ...

% This highlights an important challenge for future research: empowering LLMs not just to ``fix” RTL code, but to reason about the design intent and generate entirely new RTL formulations when appropriate. Such global optimization requires models to reason across functional correctness, microarchitecture trade-offs, and the downstream implications on synthesis. These capabilities remain largely underexplored by current LLMs applied to generative RTL optimization.

\section{Conclusions}\label{sec:conclusion}

In this paper, we introduce RTL-OPT, a benchmark for RTL code optimization aimed at enhancing IC design quality. RTL-OPT includes 36 handcrafted digital IC designs, each with suboptimal and optimized RTL code, enabling the assessment of LLM-generated RTL. An integrated evaluation framework verifies functional correctness and quantifies PPA improvements, providing a standardized method for evaluating generative AI models in hardware design. RTL-OPT has significant potential to influence AI-assisted IC design by offering valuable insights and fostering advancements. 
% As for the limitations of RTL-OPT, it relies entirely on expert-written, manually optimized RTL code, limiting the dataset’s scale. Expanding to a larger dataset requires advances in automated optimization or synthetic generation of high-quality RTL, which remains challenging.

%% file: _txt/appendix.tex
% This appendix presents comprehensive experimental results that supplement our main analysis. We include complete synthesis data for RTL-OPT under DC with 0.1ns timing constraints, along with reproduced results from RTLRewriter for comparative analysis.
\newpage

\section{Full PPA Quality Comparison of RTL-OPT}
\label{sec:full-ppa}

Due to space limitations, we move the full PPA evaluation results to the appendix. This section provides the complete comparison between the suboptimal and optimized RTL designs in RTL-OPT. Both Synopsys Design Compiler and Yosys are used for synthesis and evaluation. As PPA inherently involves trade-offs, smaller values indicate better design quality. These results complement the main text by presenting full numerical evidence under different synthesis constraints.

\subsection{Results under DC Compile\_ultra, 1 ns}
\label{sec:ppa-1ns}

Table~\ref{tab:result_1} reports the full PPA metrics of all 36 RTL-OPT designs when synthesized using Synopsys DC with a relaxed 1ns clock period on the ``compile ultra'' setting. Both suboptimal and optimized RTL codes are evaluated, enabling direct comparison. This setting highlights how expert-optimized RTL consistently achieves superior power, performance, and area outcomes compared to the suboptimal versions, demonstrating the reliability of RTL-OPT for benchmarking.  

\begin{table}[H]
    % \vspace{-.1in}
    \centering
    \setlength\tabcolsep{3.5pt}
    \resizebox{0.9\textwidth}{!}{
    \renewcommand{\arraystretch}{1.1}
\begin{tabular}{c|l||ccccc|cc||ccccc|cc} \toprule
\multicolumn{2}{c||}{\multirow{3}{*}{\textbf{Design List}}} & \multicolumn{7}{c||}{\color[HTML]{BB4A0E}\textbf{Provided Suboptimal Designs}} & \multicolumn{7}{c}{\color[HTML]{2D830E}\textbf{Provided Optimized Designs}} \\[3pt] \cline{3-16} 
\multicolumn{2}{c||}{} & \multicolumn{5}{c|}{\textbf{DC Results(\textit{compile\_ultra, 1ns})}} & \multicolumn{2}{c||}{\textbf{Yosys   Results}} & \multicolumn{5}{c|}{\textbf{DC Results(\textit{compile\_ultra, 1ns})}} & \multicolumn{2}{c}{\textbf{Yosys   Results}} \\
\multicolumn{2}{c||}{} & \textbf{Cells} & \textbf{Area} & \textit{\textbf{Power}} & \textit{\textbf{WNS}} & \textit{\textbf{TNS}}  & \textbf{Cells} & \multicolumn{1}{c||}{\textbf{Area}} & \textbf{Cells} & \textbf{Area} & \textit{\textbf{Power}} & \textit{\textbf{WNS}} & \textit{\textbf{TNS}} & \textbf{Cells} & \textbf{Area} \\  \midrule\midrule
\textbf{1} & \textbf{adder} & 397 & 510.4 & 370.4 & 0.0 & 0.0  & 323 & 456.2 & 372 & 477.2 & 351.3 & 0.0 & 0.0  & 266 & 413.9 \\
% \textbf{2} & \textbf{adder\_carry} & 91 & 98.69 & 0.62 & 0.20 & 1.52 & 53 & 28 & 14.90 & 50 & 54.00 & 0.32 & 0.40 & 2.26 & 6 & 8 & 5.85 \\
\textbf{2} & \textbf{adder\_select} & 450 & 439.2 & 264.3 & 0.0 & 0.0  & 363 & 428.3 & 306 & 225.7 & 0.23 & 0.0 & 0.0 & 241 & 318.4 \\
\textbf{3} & \textbf{alu\_64bit} & 1683 & 1645 & 887.8 & 0.0 & 0.0  & 1411 & 1554  & 1680 & 1557 & 677.1 & 0.0 & 0.0 & 1094 & 1307 \\
\textbf{4} & \textbf{alu\_8bit} & 146 & 180 & 75.58 & 0.0 & 0.0  & 152 & 163.1 & 115 & 138 & 68.43 & 0.0 & 0.0  & 130 & 158.8 \\
\textbf{5} & \textbf{calculation} & 888 & 1044 & 0.78 & -0.61 & -5.66  & 774 & 900.4 & 755 & 860 & 0.62 & -0.60 & -4.83  & 567 & 661.8 \\
\textbf{6} & \textbf{comparator} & 63 & 56.39 & 0.03 & 0.0 & 0.0  & 40 & 40.70 & 45 & 41.76 & 0.02 & 0.0 & 0.0  & 40 & 40.70 \\
\textbf{7} & \textbf{comparator\_16bit} & 102  & 88.31   & 44.57 & 0.0 & 0.0  & 64 & 62.24 & 116  & 105.3  & 38.83 & 0.0 & 0.0  & 104 & 105.1 \\
\textbf{8} & \textbf{comparator\_2bit} & 10   & 9.58    & 4.51 & 0.0 & 0.0  & 11 & 9.31 & 10   & 8.78    & 3.82 & 0.0 & 0.0  & 10 & 9.31 \\
\textbf{9} & \textbf{comparator\_4bit} & 23   & 21.28   & 10.86 & 0.0 & 0.0  & 22 & 21.55 & 21   & 18.35   & 9.25 & 0.0 & 0.0  & 18 & 17.29 \\
\textbf{10} & \textbf{comparator\_8bit} & 46   & 44.42   & 22.8 & 0.0 & 0.0  & 45 & 48.15 & 47   & 41.23   & 20.29 & 0.0 & 0.0 & 33 & 30.86 \\
\textbf{11} & \textbf{decoder\_6bit}  & 97   & 76.61   & 22.74 & 0.0 & 0.0  & 93 & 99.22  & 87   & 71.55   & 19.54 & 0.0 & 0.0  & 91 & 78.20 \\
\textbf{12} & \textbf{decoder\_8bit} & 317   & 257.2   & 50.67  & 0.0 & 0.0  & 390 & 410.7 & 312   & 246.8   & 50.28 & 0.0 & 0.0  & 300 & 250 \\
\textbf{13} & \textbf{divider\_16bit} & 3471  & 3428  & 5088 & -3.43 & -75.6  & 1264 & 1412 & 1565  & 1560  & 2091 & -3.83 & -72.1 & 662 & 760.2 \\
\textbf{14} & \textbf{divider\_32bit} & 14400 & 14296 & 23796 & -10.69 & -486  & - & - & 6724  & 6703.20  & 11335 & -10.19 & -468  & 2419 & 2941 \\
\textbf{15} & \textbf{divider\_4bit} & 39    & 40.43    & 21.58 & 0.0 & 0.0  & 52 & 59.05 & 33    & 33.78    & 15.9 & 0.0 & 0.0  & 39 & 40.96 \\
\textbf{16} & \textbf{divider\_8bit} & 571   & 575.4   & 681.8 & -0.50 & -4.57  & 302 & 339.7 & 322   & 330.37   & 275.2 & -0.39 & -2.43  & 171 & 189.7 \\
\textbf{17} & \textbf{fsm} & 89    & 128.7   & 92.56  & 0.0 & 0.0  & 85 & 138.6  & 73    & 92.04    & 51.44  & 0.0 & 0.0  & 70 & 97.09 \\
\textbf{18} & \textbf{fsm\_encode} & 242   & 387.0   & 320.6 & 0.0 & 0.0  & 179 & 352.7 & 155   & 305.63   & 276.97 & 0.0 & 0.0  & 170 & 319.2 \\
\textbf{19} & \textbf{gray} & 48    & 67.30    & 85.32 & 0.0 & 0.0  & 63 & 85.92 & 51    & 69.69    & 66.78 & 0.0 & 0.0  & 67 & 88.84 \\
\textbf{20} & \textbf{mac} & 410   & 717.9   & 868.2 & 0.0 & 0.0 & 532 & 831.5 & 319   & 548.23   & 735.25 & 0.0 & 0.0  & 529 & 707 \\
\textbf{21} & \textbf{mul} & 315 & 378.5 & 421.7 & -0.05 & -0.13  & 399 & 485.5 & 315 & 378.5 & 421.7 & -0.05 & -0.13 & 397 & 476.4 \\
% \textbf{23} & \textbf{mul\_const} & 113 & 122.89 & 0.68 & 0.19 & 1.30 & 111 & 61 & 72.09 & 118 & 114.6 & 0.54 & 0.11 & 0.73 & 83 & 42 & 52.40 \\
\textbf{22} & \textbf{mul\_sub}  & 234   & 338.09   & 262.35 & 0.0 & 0.0  & 299 & 352.9 & 233   & 337.02   & 256.53 & 0.0 & 0.0  & 289 & 332.5 \\
% \textbf{25} & \textbf{mux\_4to1\_16bit} & 73 & 77.41 & 0.34 & 0.05 & 0.86 & 168 & 84 & 83.79 & 67 & 73.68 & 0.33 & 0.06 & 1.00 & 187 & 84 & 83.79 \\
% \textbf{26} & \textbf{mux\_4to1\_64bit} & 239 & 263.87 & 1.25 & 0.11 & 6.94 & 600 & 324 & 326.4 & 232 & 260.7 & 1.22 & 0.10 & 6.32 & 667 & 324 & 326.4 \\
\textbf{23} & \textbf{mux} & 25    & 31.92    & 10.38  & 0.0 & 0.0  & 8 & 8.51 & 25    & 21.81    & 8.38  & 0.0 & 0.0 & 34 & 42.29 \\
\textbf{24} & \textbf{mux\_encode} & 125 & 140.7 & 0.43 & 0.0 & 0.0  & - & - & 34 & 36.18 & 0.13 & 0.0 & 0.0  & - & - \\
\textbf{25} & \textbf{saturating\_add} & 24   & 69.43   & 67.53 & 0.0 & 0.0  & 58 & 97.36 & 18   & 67.56   & 65.77 & 0.0 & 0.0  & 42 & 78.47 \\
\textbf{26} & \textbf{selector} & 18   & 39.37   & 36.06 & 0.0 & 0.0  & 17 & 42.56 & 18   & 38.04   & 35.32 & 0.0 & 0.0  & 15 & 37.24 \\
\textbf{27} & \textbf{sub\_16bit} & 132  & 136.2  & 77.92 & 0.0 & 0.0  & 93 & 96.56 & 124  & 131.9  & 75.46  & 0.0 & 0.0  & 92 & 98.95 \\
\textbf{28} & \textbf{sub\_32bit} & 270  & 251.9  & 148.9 & 0.0 & 0.0  & 189 & 191.5 & 265  & 244.7  & 142.52 & 0.0 & 0.0 & 188 & 200 \\
\textbf{29} & \textbf{sub\_4bit} & 12 & 18.35 & 9.29 & 0.0 & 0.0  & 21 & 22.34 & 10 & 17.82 & 9.29 & 0.0 & 0.0  & 20 & 22.08 \\
\textbf{30} & \textbf{sub\_8bit} & 25   & 41.76   & 24.96 & 0.0 & 0.0  & 45 & 47.88 & 20   & 36.97   & 19.26  & 0.0 & 0.0  & 46 & 48.15\\ 
\textbf{31} & \textbf{add\_sub}        & 164  & 183.27  & 118.4 & 0.0 & 0.0  & 155 & 164.4 & 124  & 130.34  & 101.7  & 0.0 & 0.0  & 179 & 202.7 \\
\textbf{32} & \textbf{addr\_calcu}     & 78   & 131.40  & 96.07 & -0.03 & -0.06  & 197 & 222.9 & 82   & 125.55  & 90.27   & -0.01 & -0.01  & 101 & 124.7 \\
\textbf{33} & \textbf{mult\_if}        & 10  & 10.91  & 3.53 & 0.0 & 0.0   & 12  & 10.9  & 11  & 10.11  & 4.13 & 0.0 & 0.0   & 12  & 10.91 \\
\textbf{34} & \textbf{mux\_large}      & 81   & 97.62   & 48.27  & 0.0 & 0.0  & 65  & 90.17 & 81   & 96.82   & 40.84  & 0.0 & 0.0  & 112 & 120.5 \\
\textbf{35} & \textbf{register} & 3731 & 8745 & 7712 & 0.0 & 0.0  & 4500 & 9735 & 3720 & 8744 & 7708  & 0.0 & 0.0  & 4507 & 9668 \\
\textbf{36} & \textbf{ticket\_machine} & 36   & 58.52   & 60.31 & 0.0 & 0.0  & 29  & 47.88 & 22   & 32.45   & 36.03  & 0 & 0 & 29  & 47.88\\
\bottomrule
\end{tabular}
}
    \vspace{.1in}
\caption{The PPA quality comparison of RTL-OPT-provided suboptimal vs. optimized designs(compile\_ultra, 1ns). 
Using both commercial DC and open-source Yosys for the RTL design synthesis and PPA evaluations.  Trade-offs are common in these PPA metrics and smaller values indicate better performance.
}
\label{tab:result_1}
    % \vspace{-.15in}
\end{table}

\subsection{Results under DC Compile, 0.1 ns}
\label{sec:ppa-01ns}

Table~\ref{tab:result_2} presents the full PPA metrics for the same 36 designs under a more aggressive 0.1ns clock period constraint. Compared to the 1ns setting, these results illustrate sharper trade-offs among PPA metrics, where aggressive timing optimization can sometimes increase power or area. Nevertheless, the optimized RTL consistently outperforms the suboptimal RTL, reaffirming that RTL-OPT reflects realistic and meaningful optimization challenges.  

% \subsection{PPA Quality Comparison of RTL-OPT (DC Compile, 0.1 ns)}

% Table~\ref{tab:result_2} presents the full PPA metrics for all 36 designs in the RTL-OPT benchmark when synthesized using Synopsys Design Compiler under aggressive timing constraints (0.1ns clock period). In contrast to Table~\ref{tab:result_1} in the main body, which focuses on the ``compile ultra'' setting with a relaxed 1ns clock period, this table highlights the trade-offs and variations under tighter constraints. Both suboptimal and optimized designs provided by RTL-OPT are evaluated.

\begin{table}[H]
    % \vspace{-.1in}
    \centering
    \setlength\tabcolsep{3.5pt}
    \resizebox{0.9\textwidth}{!}{
    \renewcommand{\arraystretch}{1.1}
\begin{tabular}{c|l||ccccc|ccc||ccccc|ccc} \toprule
\multicolumn{2}{c||}{\multirow{3}{*}{\textbf{Design List}}} & \multicolumn{8}{c||}{\color[HTML]{BB4A0E}\textbf{Provided Suboptimal Designs}} & \multicolumn{8}{c}{\color[HTML]{2D830E}\textbf{Provided Optimized Designs}} \\[3pt] \cline{3-18} 
\multicolumn{2}{c||}{} & \multicolumn{5}{c|}{\textbf{DC Results (\textit{compile, 0.1ns})}} & \multicolumn{3}{c||}{\textbf{Yosys   Results}} & \multicolumn{5}{c|}{\textbf{DC Results (\textit{compile, 0.1ns}) }} & \multicolumn{3}{c}{\textbf{Yosys   Results}} \\
\multicolumn{2}{c||}{} & \textbf{Cells} & \textbf{Area} & \textit{\textbf{Power}} & \textit{\textbf{WNS}} & \textit{\textbf{TNS}} & \textbf{Wires} & \textbf{Cells} & \multicolumn{1}{c||}{\textbf{Area}} & \textbf{Cells} & \textbf{Area} & \textit{\textbf{Power}} & \textit{\textbf{WNS}} & \textit{\textbf{TNS}} & \textbf{Wires} & \textbf{Cells} & \textbf{Area} \\  \midrule\midrule
\textbf{1} & \textbf{adder} & 626 & 688.4 & 5.1 & 0.57 & 16.0 & 501 & 323 & 456.2 & 531 & 639.7 & 4.26 & 0.46 & 13.2 & 440 & 266 & 413.9 \\
% \textbf{2} & \textbf{adder\_carry} & 91 & 98.69 & 0.62 & 0.20 & 1.52 & 53 & 28 & 14.90 & 50 & 54.00 & 0.32 & 0.40 & 2.26 & 6 & 8 & 5.85 \\
\textbf{2} & \textbf{adder\_select} & 813 & 812.1 & 4.85 & 0.39 & 10.8 & 724 & 363 & 428.3 & 514 & 522.6 & 3.49 & 0.42 & 12.0 & 490 & 241 & 318.4 \\
\textbf{3} & \textbf{alu\_64bit} & 3248 & 3028 & 16.95 & 0.51 & 29.7 & 2358 & 1411 & 1554 & 1706 & 1748.9 & 8.58 & 0.74 & 42.7 & 1791 & 1094 & 1307 \\
\textbf{4} & \textbf{alu\_8bit} & 402 & 370 & 1.89 & 0.26 & 1.91 & 257 & 152 & 163.1 & 244 & 245.8 & 1.15 & 0.40 & 2.92 & 229 & 130 & 158.8 \\
\textbf{5} & \textbf{calculation} & 670 & 997.5 & 6.59 & 3.22 & 42.6 & 1210 & 774 & 900.4 & 534 & 761.8 & 5.25 & 3.24 & 44.3 & 927 & 567 & 661.8 \\
\textbf{6} & \textbf{comparator} & 109 & 98.15 & 0.43 & 0.19 & 0.19 & 111 & 40 & 40.70 & 90 & 80.60 & 0.34 & 0.21 & 0.21 & 108 & 40 & 40.70 \\
\textbf{7} & \textbf{comparator\_16bit} & 296 & 244.4 & 1.06 & 0.18 & 0.52 & 141 & 64 & 62.24 & 151 & 169.2 & 0.75 & 0.19 & 0.52 & 357 & 104 & 105.1 \\
\textbf{8} & \textbf{comparator\_2bit} & 14 & 14.36 & 0.06 & 0.02 & 0.05 & 40 & 11 & 9.31 & 16 & 13.03 & 0.05 & 0.01 & 0.02 & 28 & 10 & 9.31 \\
\textbf{9} & \textbf{comparator\_4bit} & 25 & 25.80 & 0.11 & 0.06 & 0.15 & 83 & 22 & 21.55 & 29 & 23.14 & 0.09 & 0.10 & 0.22 & 45 & 18 & 17.29 \\
\textbf{10} & \textbf{comparator\_8bit} & 73 & 71.29 & 0.31 & 0.12 & 0.33 & 170 & 45 & 48.15 & 81 & 67.30 & 0.28 & 0.12 & 0.34 & 77 & 33 & 30.86 \\
\textbf{11} & \textbf{decoder\_6bit} & 204 & 208 & 0.37 & 0.18 & 9.22 & 106 & 93 & 99.22 & 132 & 106.7 & 0.32 & 0.04 & 2.53 & 195 & 91 & 78.20 \\
\textbf{12} & \textbf{decoder\_8bit} & 781 & 855.9 & 1.28 & 0.27 & 58.8 & 407 & 390 & 410.7 & 435 & 373.2 & 0.83 & 0.09 & 20.8 & 617 & 300 & 250 \\
\textbf{13} & \textbf{divider\_16bit} & 5037 & 5045 & 66.3 & 3.88 & 89.4 & 2385 & 1264 & 1412 & 2543 & 2426 & 26.71 & 4.10 & 90.4 & 2116 & 662 & 760.2 \\
\textbf{14} & \textbf{divider\_32bit} & 21348 & 19849 & 275.51 & 12.26 & 586 & - & - & - & 16434 & 16053 & 164 & 13.68 & 875 & 8341 & 2419 & 2941 \\
\textbf{15} & \textbf{divider\_4bit} & 85 & 84.85 & 0.48 & 0.32 & 1.29 & 112 & 52 & 59.05 & 64 & 56.13 & 0.23 & 0.17 & 0.59 & 133 & 39 & 40.96 \\
\textbf{16} & \textbf{divider\_8bit} & 744 & 731.8 & 6.94 & 1.21 & 13.5 & 539 & 302 & 339.7 & 557 & 535.5 & 3.67 & 1.10 & 10.8 & 527 & 171 & 189.7 \\
\textbf{17} & \textbf{fsm} & 149 & 192.9 & 1.06 & 0.28 & 4.40 & 209 & 85 & 138.6 & 106 & 129.8 & 0.59 & 0.24 & 2.70 & 157 & 70 & 97.09 \\
\textbf{18} & \textbf{fsm\_encode} & 353 & 488.9 & 3.6 & 0.41 & 10.8 & 380 & 179 & 352.7 & 334 & 426.7 & 3.07 & 0.39 & 8.11 & 342 & 170 & 319.2 \\
\textbf{19} & \textbf{gray} & 100 & 109.1 & 0.92 & 0.28 & 2.04 & 142 & 63 & 85.92 & 81 & 94.70 & 1.04 & 0.23 & 2.34 & 145 & 67 & 88.84 \\
\textbf{20} & \textbf{mac} & 563 & 880.7 & 8.46 & 1.23 & 20.9 & 737 & 532 & 831.5 & 477 & 701.4 & 7.26 & 1.38 & 16.2 & 699 & 529 & 707 \\
\textbf{21} & \textbf{mul} & 311 & 453 & 3.64 & 1.26 & 9.69 & 498 & 399 & 485.5 & 270 & 426.4 & 3.43 & 1.25 & 9.62 & 494 & 397 & 476.4 \\
% \textbf{23} & \textbf{mul\_const} & 113 & 122.89 & 0.68 & 0.19 & 1.30 & 111 & 61 & 72.09 & 118 & 114.6 & 0.54 & 0.11 & 0.73 & 83 & 42 & 52.40 \\
\textbf{22} & \textbf{mul\_sub} & 634 & 623.24 & 4.17 & 0.75 & 8.08 & 494 & 299 & 352.9 & 614 & 603.8 & 4.12 & 0.71 & 8.11 & 480 & 289 & 332.5 \\
% \textbf{25} & \textbf{mux\_4to1\_16bit} & 73 & 77.41 & 0.34 & 0.05 & 0.86 & 168 & 84 & 83.79 & 67 & 73.68 & 0.33 & 0.06 & 1.00 & 187 & 84 & 83.79 \\
% \textbf{26} & \textbf{mux\_4to1\_64bit} & 239 & 263.87 & 1.25 & 0.11 & 6.94 & 600 & 324 & 326.4 & 232 & 260.7 & 1.22 & 0.10 & 6.32 & 667 & 324 & 326.4 \\
\textbf{23} & \textbf{mux} & 40 & 38.30 & 0.11 & 0.01 & 0.07 & 27 & 8 & 8.51 & 25 & 31.92 & 0.1 & 0.02 & 0.14 & 68 & 34 & 42.29 \\
\textbf{24} & \textbf{mux\_encode} & 125 & 140.71 & 0.43 & 0.07 & 0.59 & - & - & - & 34 & 36.18 & 0.13 & 0.08 & 0.58 & - & - & - \\
\textbf{25} & \textbf{saturating\_add} & 159 & 176.6 & 1.31 & 0.27 & 2.25 & 122 & 58 & 97.36 & 127 & 140.9 & 1.08 & 0.36 & 2.86 & 100 & 42 & 78.47 \\
\textbf{26} & \textbf{selector} & 38 & 56.4 & 0.44 & 0.11 & 0.38 & 42 & 17 & 42.56 & 31 & 49.74 & 0.39 & 0.08 & 0.24 & 29 & 15 & 37.24 \\
\textbf{27} & \textbf{sub\_16bit} & 234 & 223.2 & 1.31 & 0.27 & 3.48 & 220 & 93 & 96.56 & 302 & 263.1 & 1.39 & 0.30 & 3.94 & 181 & 92 & 98.95 \\
\textbf{28} & \textbf{sub\_32bit} & 542 & 502.2 & 2.94 & 0.31 & 8.49 & 444 & 189 & 191.5 & 518 & 452.2 & 2.51 & 0.36 & 9.22 & 370 & 188 & 200 \\
\textbf{29} & \textbf{sub\_4bit} & 37 & 36.71 & 0.17 & 0.10 & 0.35 & 52 & 21 & 22.34 & 31 & 29.79 & 0.15 & 0.13 & 0.40 & 43 & 20 & 22.08 \\
\textbf{30} & \textbf{sub\_8bit} & 135 & 122.6 & 0.65 & 0.15 & 0.95 & 108 & 45 & 47.88 & 129 & 111.9 & 0.57 & 0.26 & 1.47 & 90 & 46 & 48.15\\ 
\textbf{31} & \textbf{add\_sub}        & 496 & 444.2 & 2.57  & 0.31 & 4.20 & 277 & 155 & 164.4 & 387 & 356.9 & 2.13  & 0.37 & 4.87 & 323 & 179 & 202.7 \\
\textbf{32} & \textbf{addr\_calcu}     & 405 & 388.1 & 3.07  & 0.62 & 8.56 & 340 & 197 & 222.9 & 229 & 214.4 & 1.8   & 0.60 & 8.47 & 192 & 101 & 124.7 \\
\textbf{33} & \textbf{mult\_if}        & 17  & 15.96  & 0.052 & 0.05 & 0.05 & 39  & 12  & 10.9  & 15  & 14.90  & 0.046 & 0.09 & 0.09 & 34  & 12  & 10.91 \\
\textbf{34} & \textbf{mux\_large}      & 296 & 273.45 & 0.84  & 0.11 & 0.85 & 210 & 65  & 90.17 & 164 & 176.62 & 0.74  & 0.13 & 1.01 & 270 & 112 & 120.5 \\
\textbf{35} & \textbf{register} & 5003 & 9780 & 79.63 & 0.40 & 245 & 8096 & 4500 & 9735 & 4481 & 9583 & 77.87 & 0.37 & 244 & 8140 & 4507 & 9668 \\
\textbf{36} & \textbf{ticket\_machine} & 53  & 74.48  & 0.77  & 0.22 & 1.79 & 73  & 29  & 47.88 & 48  & 51.34  & 0.45  & 0.17 & 0.83 & 73  & 29  & 47.88\\
\bottomrule
\end{tabular}
}
    \vspace{.1in}
\caption{The PPA quality comparison of RTL-OPT-provided suboptimal vs. optimized designs (compile, 0.1ns). Using both commercial DC and open-source Yosys for the RTL design synthesis and PPA evaluations.  Trade-offs are common in these PPA metrics and smaller values indicate better performance.}
\label{tab:result_2}
    % \vspace{-.15in}
\end{table}

% \newpage
\section{PPA and Functional Correctness of LLM-Optimized Designs}
 \label{sec:app:ppa}
Due to space limitations, the detailed experimental results are moved to the appendix. 
They report PPA quality and functional correctness for all designs optimized by \textbf{GPT-4o-mini}, \textbf{Gemini-2.5}, \textbf{Deepseek V3}, and \textbf{Deepseek R1}, using the RTL-OPT benchmark. 

Table~\ref{tab:exp} and~\ref{tab:exp5} summarize the evaluated PPA performance of each LLM-optimized design and compare it with the provided suboptimal RTL and optimized RTL (golden reference). \colorbox[HTML]{B7EFA5}{Green cells} indicate that the PPA is better than the suboptimal RTL, and \colorbox[HTML]{B7EFA5}{\textbf{bold green cells}} indicate that the PPA surpasses the optimized RTL (golden reference). The table also shows the functional correctness after verification (Func column), with \ding{52} and \colorbox[HTML]{FFCCC9}{\ding{56}} representing the verification results. \colorbox[HTML]{FFCCC9}{\ding{56}\ding{56}} indicates that the corresponding design contains syntax errors and fails to pass DC synthesis.

\begin{table}[H]
\vspace{-.35in}
    \centering
\resizebox{0.92\textwidth}{!}{
\begin{tabular}{l|cccccc|cccccc}\toprule
\multicolumn{1}{c|}{} &
  \multicolumn{6}{c|}{\textbf{GPT-4o-mini}} &
  \multicolumn{6}{c}{\textbf{Gemini-2.5}} \\
\multicolumn{1}{c|}{\multirow{-2}{*}{\textbf{Category}}} &
  \textit{\textbf{Cells}} &
  \textit{\textbf{Area}} &
  \textit{\textbf{Power (mW)}} &
  \textit{\textbf{WNS (ns)}} &
  \textit{\textbf{TNS (ns)}} &
  \textit{\textbf{Check}} &
  \textit{\textbf{Cells}} &
  \textit{\textbf{Area}} &
  \textit{\textbf{Power (mW)}} &
  \textit{\textbf{WNS (ns)}} &
  \textit{\textbf{TNS (ns)}} &
  \textit{\textbf{Check}} \\ \midrule\midrule
\textbf{adder} &
  - &
  - &
  - &
  - &
  - &
  \cellcolor[HTML]{FFCCC9}\ding{56} &
  - &
  - &
  - &
  - &
  - &
  \cellcolor[HTML]{FFCCC9}\ding{56}\ding{56} \\
% \textbf{adder\_carry} &
%   91 &
%   98.69 &
%   0.62 &
%   0.20 &
%   1.52 &
%   \ding{52} &
%   98 &
%   101.35 &
%   0.56 &
%   0.18 &
%   1.22 &
%   \ding{52} \\
\textbf{adder\_select} &
  813 &
  812.1 &
  4.85 &
  -0.39 &
  -10.8 &
  \ding{52} &
  813 &
  812.10 &
  4.85 &
  -0.39 &
  -10.8 &
  \ding{52} \\
\textbf{alu\_64bit} &
  3248 &
  3028 &
  16.95 &
  -0.51 &
  -29.7 &
  \ding{52} &
  3248 &
  3028 &
  16.95 &
  -0.51 &
  -29.7 &
  \ding{52} \\
\textbf{alu\_8bit} &
  402 &
  370 &
  1.89 &
  -0.26 &
  -1.91 &
  \ding{52} &
  402 &
  370.01 &
  1.89 &
  -0.26 &
  -1.91 &
  \ding{52} \\
\textbf{calculation} &
  - &
  - &
  - &
  - &
  - &
  \cellcolor[HTML]{FFCCC9}\ding{56} &
  670 &
  997.50 &
  6.59 &
  -3.22 &
  -42.6 &
  \ding{52} \\
\textbf{comparator} &
  \cellcolor[HTML]{B7EFA5}96 &
  \cellcolor[HTML]{B7EFA5}83.26 &
  \cellcolor[HTML]{B7EFA5}0.38 &
  \cellcolor[HTML]{B7EFA5}-0.22 &
  \cellcolor[HTML]{B7EFA5}-0.22 &
  \ding{52} &
  \cellcolor[HTML]{B7EFA5}100 &
  \cellcolor[HTML]{B7EFA5}88.05 &
  \cellcolor[HTML]{B7EFA5}0.38 &
  \cellcolor[HTML]{B7EFA5}-0.22 &
  \cellcolor[HTML]{B7EFA5}-0.22 &
  \ding{52} \\
\textbf{comparator\_16bit} &
  - &
  - &
  - &
  - &
  - &
  \cellcolor[HTML]{FFCCC9}\ding{56} &
  296 &
  244.45 &
  1.06 &
  -0.18 &
  -0.52 &
  \ding{52} \\
\textbf{comparator\_2bit} &
  14 &
  14.36 &
  0.06 &
  -0.02 &
  -0.05 &
  \ding{52} &
  \cellcolor[HTML]{B7EFA5}16 &
  \cellcolor[HTML]{B7EFA5}13.03 &
  \cellcolor[HTML]{B7EFA5}0.05 &
  \cellcolor[HTML]{B7EFA5}-0.01 &
  \cellcolor[HTML]{B7EFA5}-0.02 &
  \ding{52} \\
\textbf{comparator\_4bit} &
  33 &
  29.26 &
  0.18 &
  -0.05 &
  -0.15 &
  \ding{52} &
  \cellcolor[HTML]{B7EFA5}29 &
  \cellcolor[HTML]{B7EFA5}23.14 &
  \cellcolor[HTML]{B7EFA5}0.09 &
  \cellcolor[HTML]{B7EFA5}-0.10 &
  \cellcolor[HTML]{B7EFA5}-0.22 &
  \ding{52} \\
\textbf{comparator\_8bit} &
  - &
  - &
  - &
  - &
  - &
  \cellcolor[HTML]{FFCCC9}\ding{56} &
  79 &
  83.52 &
  0.37 &
  -0.11 &
  -0.31 &
  \ding{52} \\
\textbf{decoder\_6bit} &
  \cellcolor[HTML]{B7EFA5}\textbf{128} &
  \cellcolor[HTML]{B7EFA5}\textbf{109.9} &
  \cellcolor[HTML]{B7EFA5}\textbf{0.31} &
  \cellcolor[HTML]{B7EFA5}\textbf{-0.03} &
  \cellcolor[HTML]{B7EFA5}\textbf{-1.74} &
  \ding{52} &
  204 &
  208.01 &
  0.37 &
  -0.18 &
  -9.22 &
  \ding{52} \\
\textbf{decoder\_8bit} &
  781 &
  855.9 &
  1.28 &
  -0.27 &
  -58.8 &
  \ding{52} &
  \cellcolor[HTML]{B7EFA5}\textbf{354} &
  \cellcolor[HTML]{B7EFA5}\textbf{297.4} &
  \cellcolor[HTML]{B7EFA5}\textbf{0.66} &
  \cellcolor[HTML]{B7EFA5}\textbf{-0.11} &
  \cellcolor[HTML]{B7EFA5}\textbf{-24.8} &
  \ding{52} \\
\textbf{divider\_16bit} &
  5037 &
  5045 &
  66.25 &
  -3.88 &
  -89.4 &
  \ding{52} &
  \cellcolor[HTML]{B7EFA5}3926 &
  \cellcolor[HTML]{B7EFA5}3895 &
  \cellcolor[HTML]{B7EFA5}51.75 &
  \cellcolor[HTML]{B7EFA5}-6.85 &
  \cellcolor[HTML]{B7EFA5}-155 &
  \ding{52} \\
\textbf{divider\_32bit} &
  16638 &
  16211 &
  171.11 &
  -13.23 &
  -846.3 &
  \ding{52} &
  14529 &
  13375 &
  154.6 &
  -13.24 &
  -629 &
  \ding{52} \\
\textbf{divider\_4bit} &
  85 &
  84.85 &
  0.48 &
  -0.32 &
  -1.29 &
  \ding{52} &
  - &
  - &
  - &
  - &
  - &
  \cellcolor[HTML]{FFCCC9}\ding{56} \\
\textbf{divider\_8bit} &
  744 &
  731.8 &
  6.94 &
  -1.21 &
  -13.5 &
  \ding{52} &
  \cellcolor[HTML]{B7EFA5}\textbf{531} &
  \cellcolor[HTML]{B7EFA5}\textbf{513.38} &
  \cellcolor[HTML]{B7EFA5}\textbf{3.58} &
  \cellcolor[HTML]{B7EFA5}\textbf{-1.15} &
  \cellcolor[HTML]{B7EFA5}\textbf{-11.0} &
  \ding{52} \\
\textbf{fsm} &
  149 &
  192.8 &
  1.06 &
  -0.28 &
  -4.40 &
  \ding{52} &
  - &
  - &
  - &
  - &
  - &
  \cellcolor[HTML]{FFCCC9}\ding{56}\ding{56} \\
\textbf{fsm\_encode} &
  389 &
  504.9 &
  3.69 &
  -0.41 &
  -12.5 &
  \ding{52} &
  \cellcolor[HTML]{B7EFA5}357 &
  \cellcolor[HTML]{B7EFA5}448.21 &
  \cellcolor[HTML]{B7EFA5}3.16 &
  \cellcolor[HTML]{B7EFA5}-0.40 &
  \cellcolor[HTML]{B7EFA5}-7.95 &
  \ding{52} \\
\textbf{gray} &
  - &
  - &
  - &
  - &
  - &
  \cellcolor[HTML]{FFCCC9}\ding{56} &
  100 &
  109 &
  0.92 &
  -0.28 &
  -2.04 &
  \ding{52} \\
\textbf{mac} &
  - &
  - &
  - &
  - &
  - &
  \cellcolor[HTML]{FFCCC9}\ding{56} &
  - &
  - &
  - &
  - &
  - &
  \cellcolor[HTML]{FFCCC9}\ding{56} \\
\textbf{mul} &
  311 &
  453 &
  3.64 &
  -1.26 &
  -9.69 &
  \ding{52} &
  311 &
  453 &
  3.64 &
  -1.26 &
  -9.69 &
  \ding{52} \\
% \textbf{mul\_const} &
%   117 &
%   111.2 &
%   0.49 &
%   0.11 &
%   0.65 &
%   \cellcolor[HTML]{FFCCC9}\ding{56} &
%   113 &
%   122.89 &
%   0.68 &
%   0.19 &
%   1.30 &
%   \ding{52} \\
\textbf{mul\_sub} &
  634 &
  623.2 &
  4.17 &
  -0.75 &
  -8.08 &
  \ding{52} &
  \cellcolor[HTML]{B7EFA5}282 &
  \cellcolor[HTML]{B7EFA5}381.71 &
  \cellcolor[HTML]{B7EFA5}2.67 &
  \cellcolor[HTML]{B7EFA5}-1.11 &
  \cellcolor[HTML]{B7EFA5}-11.6 &
  \ding{52} \\
% \textbf{mux\_4to1\_16bit} &
%   73 &
%   77.41 &
%   0.34 &
%   0.05 &
%   0.86 &
%   \ding{52} &
%   73 &
%   77.41 &
%   0.34 &
%   0.05 &
%   0.86 &
%   \ding{52} \\
% \textbf{mux\_4to1\_64bit} &
%   239 &
%   263.9 &
%   1.25 &
%   0.11 &
%   6.94 &
%   \ding{52} &
%   239 &
%   263.9 &
%   1.25 &
%   0.11 &
%   6.94 &
%   \ding{52} \\
\textbf{mux} &
  \cellcolor[HTML]{B7EFA5}32 &
  \cellcolor[HTML]{B7EFA5}25.54 &
  \cellcolor[HTML]{B7EFA5}0.08 &
  \cellcolor[HTML]{B7EFA5}0.0 &
  \cellcolor[HTML]{B7EFA5}0.0 &
  \ding{52} &
  \cellcolor[HTML]{B7EFA5}32 &
  \cellcolor[HTML]{B7EFA5}25.54 &
  \cellcolor[HTML]{B7EFA5}0.08 &
  \cellcolor[HTML]{B7EFA5}0.0 &
  \cellcolor[HTML]{B7EFA5}0.0 &
  \ding{52} \\
\textbf{mux\_encode} &
  - &
  - &
  - &
  - &
  - &
  \cellcolor[HTML]{FFCCC9}\ding{56} &
  - &
  - &
  - &
  - &
  - &
  \cellcolor[HTML]{FFCCC9}\ding{56} \\
\textbf{saturating\_add} &
  - &
  - &
  - &
  - &
  - &
  \cellcolor[HTML]{FFCCC9}\ding{56}\ding{56} &
  \cellcolor[HTML]{B7EFA5}127 &
  \cellcolor[HTML]{B7EFA5}140.98 &
  \cellcolor[HTML]{B7EFA5}1.08 &
  \cellcolor[HTML]{B7EFA5}-0.36 &
  \cellcolor[HTML]{B7EFA5}-2.86 &
  \ding{52} \\
\textbf{selector} &
  \cellcolor[HTML]{B7EFA5}30 &
  \cellcolor[HTML]{B7EFA5}53.73 &
  \cellcolor[HTML]{B7EFA5}0.48 &
  \cellcolor[HTML]{B7EFA5}-0.18 &
  \cellcolor[HTML]{B7EFA5}-0.49 &
  \ding{52} &
  38 &
  56.39 &
  0.44 &
  -0.11 &
  -0.38 &
  \ding{52} \\
\textbf{sub\_16bit} &
  266 &
  230.6 &
  1.22 &
  -0.27 &
  -3.40 &
  \ding{52} &
  266 &
  230.62 &
  1.22 &
  -0.27 &
  -3.40 &
  \ding{52} \\
\textbf{sub\_32bit} &
  \cellcolor[HTML]{B7EFA5}551 &
  \cellcolor[HTML]{B7EFA5}476.4 &
  \cellcolor[HTML]{B7EFA5}2.57 &
  \cellcolor[HTML]{B7EFA5}-0.35 &
  \cellcolor[HTML]{B7EFA5}-9.13 &
  \ding{52} &
  \cellcolor[HTML]{B7EFA5}551 &
  \cellcolor[HTML]{B7EFA5}476.41 &
  \cellcolor[HTML]{B7EFA5}2.57 &
  \cellcolor[HTML]{B7EFA5}-0.35 &
  \cellcolor[HTML]{B7EFA5}-9.13 &
  \ding{52} \\
\textbf{sub\_4bit} &
  50 &
  49.48 &
  0.23 &
  -0.16 &
  -0.59 &
  \ding{52} &
  - &
  - &
  - &
  - &
  - &
  \cellcolor[HTML]{FFCCC9}\ding{56} \\
\textbf{sub\_8bit} &
  \cellcolor[HTML]{B7EFA5}\textbf{121} &
  \cellcolor[HTML]{B7EFA5}\textbf{102.7} &
  \cellcolor[HTML]{B7EFA5}\textbf{0.51} &
  \cellcolor[HTML]{B7EFA5}\textbf{-0.23} &
  \cellcolor[HTML]{B7EFA5}\textbf{-1.30} &
  \ding{52} &
  - &
  - &
  - &
  - &
  - &
  \cellcolor[HTML]{FFCCC9}\ding{56} \\ 
  \textbf{add\_sub} &
  496 &
  444.22 &
  2.57 &
  -0.31 &
  -4.20 &
  \ding{52} &
  496 &
  444.22 &
  2.57 &
  -0.31 &
  -4.20 &
  \ding{52} \\
\textbf{addr\_calcu} &
  405 &
  388.09 &
  3.07 &
  -0.62 &
  -8.56 &
  \ding{52} &
  - &
  - &
  - &
  - &
  - &
  \cellcolor[HTML]{FFCCC9}\ding{56} \\
\textbf{mult\_if} &
  17 &
  15.96 &
  0.05 &
  -0.05 &
  -0.05 &
  \ding{52} &
  17 &
  15.96 &
  0.05 &
  -0.05 &
  -0.05 &
  \ding{52} \\
\textbf{mux\_large} &
  296 &
  273.45 &
  0.84 &
  -0.11 &
  -0.85 &
  \ding{52} &
  - &
  - &
  - &
  - &
  - &
  \cellcolor[HTML]{FFCCC9}\ding{56} \\
\textbf{register} &
  \cellcolor[HTML]{B7EFA5}4615 &
  \cellcolor[HTML]{B7EFA5}9480.24 &
  \cellcolor[HTML]{B7EFA5}78.21 &
  \cellcolor[HTML]{B7EFA5}-0.36 &
  \cellcolor[HTML]{B7EFA5}-251 &
  \ding{52} &
  - &
  - &
  - &
  - &
  - &
  \cellcolor[HTML]{FFCCC9}\ding{56}\ding{56} \\
\textbf{ticket\_machine} &
  - &
  - &
  - &
  - &
  - &
   \cellcolor[HTML]{FFCCC9}\ding{56} &
  - &
  - &
  - &
  - &
  - &  \cellcolor[HTML]{FFCCC9}\ding{56} \\
  % \bottomrule
  \end{tabular}
}

\resizebox{0.92\textwidth}{!}{
\begin{tabular}{l|cccccc|cccccc}\toprule
\multicolumn{1}{c|}{} &
  \multicolumn{6}{c|}{\textbf{Deepseek V3}} &
  \multicolumn{6}{c}{\textbf{Deepseek R1}} \\
\multicolumn{1}{c|}{\multirow{-2}{*}{\textbf{Category}}} &
  \textit{\textbf{Cells}} &
  \textit{\textbf{Area}} &
  \textit{\textbf{Power (mW)}} &
  \textit{\textbf{WNS (ns)}} &
  \textit{\textbf{TNS (ns)}} &
  \textit{\textbf{Check}} &
  \textit{\textbf{Cells}} &
  \textit{\textbf{Area}} &
  \textit{\textbf{Power (mW)}} &
  \textit{\textbf{WNS (ns)}} &
  \textit{\textbf{TNS (ns)}} &
  \textit{\textbf{Check}} \\ \midrule\midrule
\textbf{adder} &
  \cellcolor[HTML]{B7EFA5}\textbf{530} &
  \cellcolor[HTML]{B7EFA5}\textbf{618.7} &
  \cellcolor[HTML]{B7EFA5}\textbf{4.51} &
  \cellcolor[HTML]{B7EFA5}\textbf{-0.43} &
  \cellcolor[HTML]{B7EFA5}\textbf{-12.5} &
  \ding{52} &
  \cellcolor[HTML]{B7EFA5}597 &
  \cellcolor[HTML]{B7EFA5}669.8 &
  \cellcolor[HTML]{B7EFA5}4.78 &
  \cellcolor[HTML]{B7EFA5}-0.41 &
  \cellcolor[HTML]{B7EFA5}-11.8 &
  \ding{52} \\
% \textbf{adder\_carry} &
%   98 &
%   101.4 &
%   0.56 &
%   0.18 &
%   1.22 &
%   \ding{52} &
%   98 &
%   101.4 &
%   0.56 &
%   0.18 &
%   1.22 &
%   \ding{52} \\
\textbf{adder\_select} &
  813 &
  812.1 &
  4.85 &
  -0.39 &
  -10.8 &
  \ding{52} &
  \cellcolor[HTML]{B7EFA5}514 &
  \cellcolor[HTML]{B7EFA5}522.7 &
  \cellcolor[HTML]{B7EFA5}3.49 &
  \cellcolor[HTML]{B7EFA5}-0.42 &
  \cellcolor[HTML]{B7EFA5}-12.0 &
  \ding{52} \\
\textbf{alu\_64bit} &
  3248 &
  3028 &
  16.95 &
  -0.51 &
  -29.7 &
  \ding{52} &
  3248 &
  3028 &
  16.95 &
  -0.51 &
  -29.7 &
  \ding{52} \\
\textbf{alu\_8bit} &
  402 &
  370 &
  1.89 &
  -0.26 &
  -1.91 &
  \ding{52} &
  402 &
  370 &
  1.89 &
  -0.26 &
  -1.91 &
  \cellcolor[HTML]{FFCCC9}\ding{56} \\
\textbf{calculation} &
  670 &
  997.5 &
  6.59 &
  -3.22 &
  -42.6 &
  \ding{52} &
  \cellcolor[HTML]{B7EFA5}550 &
  \cellcolor[HTML]{B7EFA5}808.1 &
  \cellcolor[HTML]{B7EFA5}5.73 &
  \cellcolor[HTML]{B7EFA5}3.29 &
  \cellcolor[HTML]{B7EFA5}44.8 &
  \ding{52} \\
\textbf{comparator} &
  - &
  - &
  - &
  - &
  - &
  \cellcolor[HTML]{FFCCC9}\ding{56} &
  110 &
  96.29 &
  0.4 &
  -0.19 &
  -0.19 &
  \ding{52} \\
\textbf{comparator\_16bit} &
  - &
  - &
  - &
  - &
  - &
  \cellcolor[HTML]{FFCCC9}\ding{56} &
  \cellcolor[HTML]{B7EFA5}\textbf{122} &
  \cellcolor[HTML]{B7EFA5}\textbf{107.2} &
  \cellcolor[HTML]{B7EFA5}\textbf{0.48} &
  \cellcolor[HTML]{B7EFA5}\textbf{-0.17} &
  \cellcolor[HTML]{B7EFA5}\textbf{-0.48} &
  \ding{52} \\
\textbf{comparator\_2bit} &
  \cellcolor[HTML]{B7EFA5}\textbf{10} &
  \cellcolor[HTML]{B7EFA5}\textbf{9.84} &
  \cellcolor[HTML]{B7EFA5}\textbf{0.04} &
  \cellcolor[HTML]{B7EFA5}\textbf{-0.03} &
  \cellcolor[HTML]{B7EFA5}\textbf{-0.03} &
  \ding{52} &
  \cellcolor[HTML]{B7EFA5}\textbf{13} &
  \cellcolor[HTML]{B7EFA5}\textbf{11.17} &
  \cellcolor[HTML]{B7EFA5}\textbf{0.04} &
  \cellcolor[HTML]{B7EFA5}\textbf{-0.03} &
  \cellcolor[HTML]{B7EFA5}\textbf{-0.04} &
  \ding{52} \\
\textbf{comparator\_4bit} &
  \cellcolor[HTML]{B7EFA5}\textbf{25} &
  \cellcolor[HTML]{B7EFA5}\textbf{25.80} &
  \cellcolor[HTML]{B7EFA5}\textbf{0.11} &
  \cellcolor[HTML]{B7EFA5}\textbf{-0.06} &
  \cellcolor[HTML]{B7EFA5}\textbf{-0.15} &
  \ding{52} &
  - &
  - &
  - &
  - &
  - &
  \cellcolor[HTML]{FFCCC9}\ding{56}\ding{56} \\
\textbf{comparator\_8bit} &
  73 &
  71.29 &
  0.31 &
  -0.12 &
  -0.33 &
  \ding{52} &
  - &
  - &
  - &
  - &
  - &
  \cellcolor[HTML]{FFCCC9}\ding{56}\ding{56} \\
\textbf{decoder\_6bit} &
  204 &
  208.01 &
  0.37 &
  -0.18 &
  -9.22 &
  \ding{52} &
  - &
  - &
  - &
  - &
  - &
  \cellcolor[HTML]{FFCCC9}\ding{56} \\
\textbf{decoder\_8bit} &
  781 &
  856 &
  1.28 &
  -0.27 &
  -58.8 &
  \ding{52} &
  \cellcolor[HTML]{B7EFA5}\textbf{392} &
  \cellcolor[HTML]{B7EFA5}\textbf{330.1} &
  \cellcolor[HTML]{B7EFA5}\textbf{0.72} &
  \cellcolor[HTML]{B7EFA5}\textbf{-0.10} &
  \cellcolor[HTML]{B7EFA5}\textbf{-23.1} &
  \ding{52} \\
\textbf{divider\_16bit} &
  - &
  - &
  - &
  - &
  - &
  \cellcolor[HTML]{FFCCC9}\ding{56} &
  5037 &
  5045 &
  66.25 &
  -3.88 &
  -89.4 &
  \ding{52} \\
\textbf{divider\_32bit} &
  - &
  - &
  - &
  - &
  - &
  \cellcolor[HTML]{FFCCC9}\ding{56} &
  - &
  - &
  - &
  - &
  - &
  \cellcolor[HTML]{FFCCC9}\ding{56}\ding{56} \\
\textbf{divider\_4bit} &
  - &
  - &
  - &
  - &
  - &
  \cellcolor[HTML]{FFCCC9}\ding{56} &
  - &
  - &
  - &
  - &
  - &
  \cellcolor[HTML]{FFCCC9}\ding{56} \\
\textbf{divider\_8bit} &
  - &
  - &
  - &
  - &
  - &
  \cellcolor[HTML]{FFCCC9}\ding{56} &
  - &
  - &
  - &
  - &
  - &
  \cellcolor[HTML]{FFCCC9}\ding{56}\ding{56} \\
\textbf{fsm} &
  \cellcolor[HTML]{B7EFA5}\textbf{100} &
  \cellcolor[HTML]{B7EFA5}\textbf{129.5} &
  \cellcolor[HTML]{B7EFA5}\textbf{0.6} &
  \cellcolor[HTML]{B7EFA5}\textbf{-0.22} &
  \cellcolor[HTML]{B7EFA5}\textbf{-2.62} &
  \ding{52} &
  \cellcolor[HTML]{B7EFA5}106 &
  \cellcolor[HTML]{B7EFA5}129.8 &
  \cellcolor[HTML]{B7EFA5}0.59 &
  \cellcolor[HTML]{B7EFA5}-0.24 &
  \cellcolor[HTML]{B7EFA5}-2.70 &
  \ding{52} \\
\textbf{fsm\_encode} &
  \cellcolor[HTML]{B7EFA5}372 &
  \cellcolor[HTML]{B7EFA5}466.8 &
  \cellcolor[HTML]{B7EFA5}3.46 &
  \cellcolor[HTML]{B7EFA5}-0.38 &
  \cellcolor[HTML]{B7EFA5}-9.36 &
  \ding{52} &
  \cellcolor[HTML]{B7EFA5}357 &
  \cellcolor[HTML]{B7EFA5}448.2 &
  \cellcolor[HTML]{B7EFA5}3.16 &
  \cellcolor[HTML]{B7EFA5}-0.40 &
  \cellcolor[HTML]{B7EFA5}-7.95 &
  \ding{52} \\
\textbf{gray} &
  100 &
  109.1 &
  0.92 &
  -0.28 &
  -2.04 &
  \ding{52} &
  100 &
  109.1 &
  0.92 &
  -0.28 &
  -2.04 &
  \ding{52} \\
\textbf{mac} &
  - &
  - &
  - &
  - &
  - &
  \cellcolor[HTML]{FFCCC9}\ding{56} &
  - &
  - &
  - &
  - &
  - &
  \cellcolor[HTML]{FFCCC9}\ding{56} \\
\textbf{mul} &
  - &
  - &
  - &
  - &
  - &
  \cellcolor[HTML]{FFCCC9}\ding{56} &
  270 &
  426.4 &
  3.43 &
  -1.25 &
  -9.62 &
  \ding{52} \\
% \textbf{mul\_const} &
%   \cellcolor[HTML]{B7EFA5}118 &
%   \cellcolor[HTML]{B7EFA5}114.6 &
%   \cellcolor[HTML]{B7EFA5}0.54 &
%   \cellcolor[HTML]{B7EFA5}0.11 &
%   \cellcolor[HTML]{B7EFA5}0.73 &
%   \ding{52} &
%   118 &
%   114.7 &
%   0.54 &
%   0.11 &
%   0.73 &
%   \ding{52} \\
\textbf{mul\_sub} &
  657 &
  654.4 &
  4.44 &
  -0.76 &
  -8.08 &
  \ding{52} &
  675 &
  654.1 &
  4.46 &
  -0.70 &
  -7.97 &
  \ding{52} \\
% \textbf{mux\_4to1\_16bit} &
%   73 &
%   77.41 &
%   0.34 &
%   0.05 &
%   0.86 &
%   \ding{52} &
%   73 &
%   77.41 &
%   0.34 &
%   0.05 &
%   0.86 &
%   \ding{52} \\
% \textbf{mux\_4to1\_64bit} &
%   239 &
%   263.9 &
%   1.25 &
%   0.11 &
%   6.94 &
%   \ding{52} &
%   239 &
%   263.9 &
%   1.25 &
%   0.11 &
%   6.94 &
%   \ding{52} \\
\textbf{mux} &
  \cellcolor[HTML]{B7EFA5}32 &
  \cellcolor[HTML]{B7EFA5}25.54 &
  \cellcolor[HTML]{B7EFA5}0.08 &
  \cellcolor[HTML]{B7EFA5}0.0 &
  \cellcolor[HTML]{B7EFA5}0.0 &
  \ding{52} &
  \cellcolor[HTML]{B7EFA5}32 &
  \cellcolor[HTML]{B7EFA5}25.54 &
  \cellcolor[HTML]{B7EFA5}0.82 &
  \cellcolor[HTML]{B7EFA5}0.0 &
  \cellcolor[HTML]{B7EFA5}0.0 &
  \ding{52} \\
\textbf{mux\_encode} &
  - &
  - &
  - &
  - &
  - &
  \cellcolor[HTML]{FFCCC9}\ding{56} &
  \cellcolor[HTML]{B7EFA5}113 &
  \cellcolor[HTML]{B7EFA5}121.6 &
  \cellcolor[HTML]{B7EFA5}0.46 &
  \cellcolor[HTML]{B7EFA5}-0.09 &
  \cellcolor[HTML]{B7EFA5}-0.69 &
  \ding{52} \\
\textbf{saturating\_add} &
  \cellcolor[HTML]{B7EFA5}127 &
  \cellcolor[HTML]{B7EFA5}140.9 &
  \cellcolor[HTML]{B7EFA5}1.08 &
  \cellcolor[HTML]{B7EFA5}-0.36 &
  \cellcolor[HTML]{B7EFA5}-2.86 &
  \ding{52} &
  - &
  - &
  - &
  - &
  - &
  \cellcolor[HTML]{FFCCC9}\ding{56} \\
\textbf{selector} &
  38 &
  56.39 &
  0.44 &
  -0.11 &
  -0.38 &
  \ding{52} &
  \cellcolor[HTML]{B7EFA5}30 &
  \cellcolor[HTML]{B7EFA5}49.74 &
  \cellcolor[HTML]{B7EFA5}0.41 &
  \cellcolor[HTML]{B7EFA5}-0.18 &
  \cellcolor[HTML]{B7EFA5}-0.49 &
  \ding{52} \\
\textbf{sub\_16bit} &
  266 &
  230.6 &
  1.22 &
  -0.27 &
  -3.40 &
  \ding{52} &
  \cellcolor[HTML]{B7EFA5}\textbf{266} &
  \cellcolor[HTML]{B7EFA5}\textbf{230.6} &
  \cellcolor[HTML]{B7EFA5}\textbf{1.22} &
  \cellcolor[HTML]{B7EFA5}\textbf{-0.27} &
  \cellcolor[HTML]{B7EFA5}\textbf{-3.40} &
  \ding{52} \\
\textbf{sub\_32bit} &
  \cellcolor[HTML]{B7EFA5}551 &
  \cellcolor[HTML]{B7EFA5}476.4 &
  \cellcolor[HTML]{B7EFA5}2.57 &
  \cellcolor[HTML]{B7EFA5}-0.35 &
  \cellcolor[HTML]{B7EFA5}-9.13 &
  \ding{52} &
  - &
  - &
  - &
  - &
  - &
  \cellcolor[HTML]{FFCCC9}\ding{56} \\
\textbf{sub\_4bit} &
  37 &
  37.24 &
  0.18 &
  -0.10 &
  -0.30 &
  \ding{52} &
  \cellcolor[HTML]{B7EFA5}37 &
  \cellcolor[HTML]{B7EFA5}37.24 &
  \cellcolor[HTML]{B7EFA5}0.12 &
  \cellcolor[HTML]{B7EFA5}-0.10 &
  \cellcolor[HTML]{B7EFA5}-0.30 &
  \ding{52} \\
\textbf{sub\_8bit} &
  \cellcolor[HTML]{B7EFA5}\textbf{121} &
  \cellcolor[HTML]{B7EFA5}\textbf{102.7} &
  \cellcolor[HTML]{B7EFA5}\textbf{0.51} &
  \cellcolor[HTML]{B7EFA5}\textbf{-0.23} &
  \cellcolor[HTML]{B7EFA5}\textbf{-1.30} &
  \ding{52} &
  \cellcolor[HTML]{B7EFA5}\textbf{121} &
  \cellcolor[HTML]{B7EFA5}\textbf{102.7} &
  \cellcolor[HTML]{B7EFA5}\textbf{0.51} &
  \cellcolor[HTML]{B7EFA5}\textbf{-0.23} &
  \cellcolor[HTML]{B7EFA5}\textbf{-1.30} &
  \ding{52}\\ 
  \textbf{add\_sub} &
  496 &
  444.2 &
  2.57 &
  -0.31 &
  -4.20 &
  \ding{52} &
  - &
  - &
  - &
  - &
  - &
  \cellcolor[HTML]{FFCCC9}\ding{56} \\
\textbf{addr\_calcu} &
  229 &
  214.40 &
  1.8 &
  -0.60 &
  -8.47 &
  \ding{52} &
  - &
  - &
  - &
  - &
  - &
  \cellcolor[HTML]{FFCCC9}\ding{56} \\
\textbf{mult\_if} &
  - &
  - &
  - &
  - &
  - &
  \cellcolor[HTML]{FFCCC9}\ding{56} &
  - &
  - &
  - &
  - &
  - &
  \cellcolor[HTML]{FFCCC9}\ding{56} \\
\textbf{mux\_large} &
  \cellcolor[HTML]{B7EFA5}173 &
  \cellcolor[HTML]{B7EFA5}174.5 &
  \cellcolor[HTML]{B7EFA5}0.62 &
  \cellcolor[HTML]{B7EFA5}-0.12 &
  \cellcolor[HTML]{B7EFA5}-0.95 &
  \ding{52} &
  \cellcolor[HTML]{B7EFA5}173 &
  \cellcolor[HTML]{B7EFA5}174.5 &
  \cellcolor[HTML]{B7EFA5}0.62 &
  \cellcolor[HTML]{B7EFA5}-0.12 &
  \cellcolor[HTML]{B7EFA5}-0.95 &
  \ding{52} \\
\textbf{register} &
  - &
  - &
  - &
  - &
  - &
  \cellcolor[HTML]{FFCCC9}\ding{56} &
  5003 &
  9780 &
  79.63 &
  -0.40 &
  -245 &
  \ding{52} \\
\textbf{ticket\_machine} &
  - &
  - &
  - &
  - &
  - &
   \cellcolor[HTML]{FFCCC9}\ding{56} &
  - &
  - &
  - &
  - &
  - & 
  \cellcolor[HTML]{FFCCC9}\ding{56}\\
  \bottomrule
\end{tabular}
}
% \vspace{-.05in}
\caption{PPA quality (DC \textbf{\textit{compile, 0.1ns}}) and functional correctness for all designs optimized by GPT-4o-mini, Gemini-2.5, Deepseek V3, and Deepseek R1, using the RTL-OPT benchmark.}
\label{tab:exp5}
\end{table}

\subsection{PPA and Functional Correctness of LLM-Optimized Designs (DC Compile, 1 ns)}

Table~\ref{tab:exp5} summarizes the PPA results and functional correctness checks for designs optimized by GPT-4o-mini, Gemini-2.5, Deepseek V3, and Deepseek R1 on the RTL-OPT benchmark. All evaluations are conducted under DC compile with 0.1ns clock period. This supplements Table~\ref{tab:exp}, which uses a more relaxed timing setup.

% \newpage
\section{Reproduction of Existing Benchmark with Multiple Synthesis Flows}
\label{sec:app:rtlrewriter}

\subsection{Reproduction of \cite{yao2024rtlrewriter} Designs with Multiple Synthesis Flows}

Table~\ref{tab:exp6} provides reproduction results for 14 designs originally optimized by \cite{yao2024rtlrewriter}. Both the baseline (original) and expert-optimized versions are evaluated using our unified flow, across three synthesis settings: Yosys, DC compile (0.1ns), and DC compile ultra (1ns). This ensures fair comparisons across benchmarks and demonstrates the effectiveness of our pipeline in capturing prior work.

% \usepackage{multirow}
\begin{table}[H]

\label{tab:rtlrewriter}
\resizebox{0.85\textwidth}{!}{
\begin{tabular}{l||cc|ccccc|ccccc}\toprule
\multirow{2}{*}{\textbf{Design List}} &
  \multicolumn{2}{c|}{\textbf{Yosys}} &
  \multicolumn{5}{c|}{\textbf{DC (compile, 0.1ns)}} &
  \multicolumn{5}{c}{\textbf{DC (compile\_ultra, 1ns)}} \\
 &
  \textit{\textbf{Cells}} &
  \textit{\textbf{Area}} &
  \textit{\textbf{Cells}} &
  \textit{\textbf{Area}} &
  \textit{\textbf{Power}} &
  \textit{\textbf{WNS}} &
  \textit{\textbf{TNS}} &
  \textit{\textbf{Cells}} &
  \textit{\textbf{Area}} &
  \textit{\textbf{Power}} &
  \textit{\textbf{WNS}} &
  \textit{\textbf{TNS}} \\ \midrule\midrule
\textbf{case1}       & 18   & 57.99    & -  & -   & -   & - & -  & -   & -   & -  & - & -  \\
\textbf{case1\_opt}  & 12   & 30.86    & -  & -   & -   & - & -  &-   & -   & -  & - & -  \\\midrule
\textbf{case2}      & 37   & 44.95    & -  & -   & -   & - & -  & -   & -   & -  & - & -  \\
\textbf{case2\_opt}  & 37   & 44.95    & -  & -   & -   & - & -  & -   & -   & -  & - & -  \\\midrule
\textbf{case3}       & 107   & 129.3    & 289  & 255.1   & 2.25 mW   & -0.41 & -3.23  & 26   & 98.95   & 74.85 uW  & 0.0 & 0.0  \\
\textbf{case3\_opt}  & 102   & 127.7    & 289  & 255.1   & 2.25 mW   & -0.41 & -3.23  & 26   & 98.95   & 74.85 uW  & 0.0 & 0.0  \\\midrule
\textbf{case4}       & 304   & 347.1    & 576  & 550.9   & 3.75 mW   & -0.51 & -3.72  & 204  & 267.06  & 169.5 uW & 0.0 & 0.0  \\
\textbf{case4\_opt}  & 262   & 310.7    & 553  & 529.9   & 3.80 mW   & -0.50 & -3.65  & 233  & 291.27  & 165.6 uW & 0.0 & 0.0  \\\midrule
\textbf{case5}       & 10933 & 12457    & 6546 & 11104    & 173.7 mW & -4.15 & -413.6 & 6047 & 9668    & 19.99 mW  & -1    & -56.61 \\
\textbf{case5\_opt}  & 10504 & 11980    & 6044 & 10469    & 168.9 mW & -4.22 & -418.1  & 5962 & 9600 & 19.98 mW  & -1.02 & -56.7  \\\midrule
\textbf{case6}       & 37    & 59.32    & 53   & 60.38    & 484.8 uW & -0.22 & -0.68  & 28   & 41.76   & 42.93 uW  & 0.0 & 0.0  \\
\textbf{case6\_opt}  & 21    & 28.99    & 23   & 28.99    & 284.1 uW & -0.15 & -0.28  & 11   & 19.95   & 23.42 uW  & 0.0 & 0.0  \\\midrule
\textbf{case7}       & 24    & 38.84    & 22   & 34.31    & 418.7 uW & -0.15 & -0.46  & 16   & 28.99   & 40.65 uW  & 0.0 & 0.0  \\
\textbf{case7\_opt}  & 11    & 19.95    & 11   & 18.35    & 215.8 uW & -0.12 & -0.25  & 8    & 15.96   & 21.90 uW  & 0.0 & 0.0  \\\midrule
\textbf{case8}       & 1471  & 1720     & 1353 & 2242     & 13.90 mW  & -0.92 & -58.2  & 703  & 831.2  & 634.7 uW & 0.0 & 0.0  \\
\textbf{case8\_opt}  & 661   & 758.1    & 1771 & 1962     & 14.87 mW  & -0.95 & -57.0  & 711  & 834.7  & 647.3 uW & 0.0 & 0.0  \\\midrule
\textbf{case9}       & 1716  & 2010     & 1336 & 2420     & 13.31 mW  & -1.05 & -65.7  & 729  & 886.3  & 636.4 uW & 0.0 & 0.0  \\
\textbf{case9\_opt}  & 776   & 926.5    & 1757 & 1984     & 16.50 mW  & -1.00 & -70.6  & 973  & 965.1  & 982.7 uW & 0.0 & 0.0  \\\midrule
\textbf{case10}       & 24    & 43.89    & 25   & 31.92    & 103.8 uW & -0.02 & -0.14  & 25   & 31.92   & 10.38 uW  & 0.0 & 0.0  \\
\textbf{case10\_opt}  & 24    & 8.51     & 40   & 38.30    & 114.8 uW & -0.01 & -0.07  & 25   & 21.81   & 8.38 uW   & 0.0 & 0.0  \\\midrule
\textbf{case11}       & 2     & 1.86     & 9    & 5.32     & 16.07 uW  & 0.0 & 0.0  & 1    & 1.06    & 252.5 nW & 0.0 & 0.0  \\
\textbf{case11\_opt}  & 1     & 1.86     & 6    & 2.93     & 7.28 uW   & 0.0 & 0.0  & 1    & 1.06    & 252.5 nW & 0.0 & 0.0  \\\midrule
\textbf{case12}      & 1     & 1.86     & 3    & 0.0      & 374.8 nW & 0.0 & 0.0  & 3    & 0.0    & 37.48 nW  & 0.0 & 0.0  \\
\textbf{case12\_opt} & 1     & 1.06     & 1    & 1.06     & 2.59 uW   & 0.0 & 0.0  & 1    & 1.06    & 258.8 nW & 0.0 & 0.0  \\\midrule
\textbf{case13}      & 3     & 1.86     & 3    & 5.59     & 16.93 uW  & -0.03 & -0.03  & 5    & 5.05    & 1.77 uW   & 0.0 & 0.0  \\
\textbf{case13\_opt} & 2     & 1.86     & 2    & 3.72     & 12.41 uW  & -0.03 & -0.03  & 5    & 5.05    & 1.77 uW   & 0.0 & 0.0  \\\midrule
\textbf{case14}      & 4     & 1.86     & 6    & 7.98     & 28.15 uW  & -0.10 & -0.10  & 5    & 5.05    & 1.77 uW   & 0.0 & 0.0  \\
\textbf{case14\_opt} & 3     & 1.86     & 6    & 6.65     & 23.20 uW  & -0.08 & -0.08  & 5    & 5.05    & 1.77 uW   & 0.0 & 0.0 \\\bottomrule
\end{tabular}
}
% \vspace{-.05in}
\caption{Reproduction PPA results of 14 designs in~\cite{yao2024rtlrewriter} (both original and expert-optimized versions) using Yosys and DC synthesis flows.}
\label{tab:exp6}
\end{table}

\subsection{Reproduction of \cite{yao2024rtlrewriter} GPT-Optimized Designs}
\label{sec:app:ours}

Table~\ref{tab:exp7} and Table~\ref{tab:exp8} provide the PPA reproduction of designs optimized by GPT-based methods and \cite{yao2024rtlrewriter} own optimization strategies, respectively. These results help evaluate the generalizability of our flow when applied to designs outside the RTL-OPT benchmark and provide fair comparisons across different optimization methodologies. The same three synthesis settings are used (Yosys, DC compile 0.1ns, and DC compile ultra 1ns), under our unified evaluation flow.

\begin{table}[t]

\label{tab:rtlrewriter_GPT}
\resizebox{0.85\textwidth}{!}{
\begin{tabular}{l||cc|ccccc|ccccc}\toprule
\multirow{2}{*}{\textbf{Design List}} &
  \multicolumn{2}{c|}{\textbf{Yosys}} &
  \multicolumn{5}{c|}{\textbf{DC (compile, 0.1ns)}} &
  \multicolumn{5}{c}{\textbf{DC (compile\_ultra, 1ns)}} \\
 &
  \textit{\textbf{Cells}} &
  \textit{\textbf{Area}} &
  \textit{\textbf{Cells}} &
  \textit{\textbf{Area}} &
  \textit{\textbf{Power}} &
  \textit{\textbf{WNS}} &
  \textit{\textbf{TNS}} &
  \textit{\textbf{Cells}} &
  \textit{\textbf{Area}} &
  \textit{\textbf{Power}} &
  \textit{\textbf{WNS}} &
  \textit{\textbf{TNS}} \\ \midrule\midrule

\textbf{case1\_GPT}  & 18 & 57.99 & 18  & 57.99 & 584.6 uW  & -0.24 & -0.24  & 18 & 57.99 & 61.13 uW & 0.00 & 0.00 \\
\textbf{case2\_GPT}  & 37 & 44.95 & -  & - & -  & - & -  & - & - & - & - &  \\
\textbf{case3\_GPT}  & 107 & 129.3 & 289  & 255.1 & 2.25 mW  & -0.41 & -3.23  & 26 & 98.95 & 74.85 uW & 0.00 & 0.00 \\
\textbf{case4\_GPT}  & 290 & 326.6 & 637  & 602.8 & 4.04 mW  & -0.47 & -3.33  & 205 & 265.5 & 159.5 uW & 0.00 & 0.00 \\
\textbf{case5\_GPT}  & 10627 & 12090 & 5987 & 10483 & 165.9 mW & -4.18 & -408   & 5962 & 9600 & 19.98 mW & -1.02 & -56.72 \\
\textbf{case6\_GPT}  & 37 & 59.32 & 50   & 59.85 & 492.7 uW & -0.25 & -0.73  & 25 & 38.57 & 40.19 uW & 0.00 & 0.00 \\
\textbf{case7\_GPT}  & 26 & 44.69 & 30   & 43.62 & 522.1 uW & -0.15 & -0.55  & 17 & 34.31 & 47.46 uW & 0.00 & 0.00 \\
\textbf{case8\_GPT}  & 844 & 952.3 & 2129 & 1915  & 15.04 mW & -0.67 & -44.17 & 704 & 932.1 & 722.7 uW & 0.00 & 0.00 \\
\textbf{case9\_GPT}  & 844 & 952.3 & 2129 & 1915  & 15.04 mW & -0.67 & -44.17 & 704 & 932.1 & 722.7 uW & 0.00 & 0.00 \\
\textbf{case10\_GPT}  & 24 & 21.38 & 32   & 25.54 & 81.96 uW & 0.0   & 0.0    & 25 & 21.81 & 8.382 uW & 0.00 & 0.00 \\
\textbf{case11\_GPT}  & 3 & 1.862 & 3    & 5.59  & 16.86 uW & -0.03 & -0.03  & 4 & 3.724 & 1.339 uW & 0.00 & 0.00 \\
\textbf{case12\_GPT} & 1 & 1.064 & 1    & 1.06  & 2.59 uW  & 0.0   & 0.0    & 1 & 1.064 & 258.8 nW & 0.00 & 0.00 \\
\textbf{case13\_GPT} & 3 & 5.586 & 6    & 5.59  & 20.58 uW & 0.0   & 0.0   & 5 & 5.054 & 1.767 uW & 0.00 & 0.00 \\
\textbf{case14\_GPT} & 2 & 3.724 & 5    & 4.26  & 16.58 uW & 0.0   & 0.0    & 4 & 3.724 & 1.248 uW & 0.00 & 0.00 \\\bottomrule
\end{tabular}
}
% \vspace{-.05in}
\caption{Reproduction PPA results of GPT-4 optimized designs in~\cite{yao2024rtlrewriter} using Yosys and DC synthesis flows.}
\label{tab:exp7}
\end{table}

\begin{table}[t]

\label{tab:rtlrewriter_rtl}
\resizebox{0.85\textwidth}{!}{
\begin{tabular}{l||cc|ccccc|ccccc}\toprule
\multirow{2}{*}{\textbf{Design List}} &
  \multicolumn{2}{c|}{\textbf{Yosys}} &
  \multicolumn{5}{c|}{\textbf{DC (compile, 0.1ns)}} &
  \multicolumn{5}{c}{\textbf{DC (compile\_ultra, 1ns)}} \\
 &
  \textit{\textbf{Cells}} &
  \textit{\textbf{Area}} &
  \textit{\textbf{Cells}} &
  \textit{\textbf{Area}} &
  \textit{\textbf{Power}} &
  \textit{\textbf{WNS}} &
  \textit{\textbf{TNS}} &
  \textit{\textbf{Cells}} &
  \textit{\textbf{Area}} &
  \textit{\textbf{Power}} &
  \textit{\textbf{WNS}} &
  \textit{\textbf{TNS}} \\ \midrule\midrule
  
\textbf{case1\_ours} & 14 & 39.9 & 13 & 40.96 & 431.3 uW & -0.21 & -0.28 & 14 & 39.90 & 44.67 uW & 0.0 & 0.0 \\
\textbf{case2\_ours} & 37 & 44.95 & 76 & 69.43 & 383.9 uW & -0.17 & -1.07 & 9 & 32.45 & 15.31 uW & 0.0 & 0.0 \\
\textbf{case3\_ours} & 107 & 129.3 & 289 & 255.1 & 2.251 mW & -0.41 & -3.23 & 26 & 98.95 & 74.85 uW & 0.0 & 0.0 \\
\textbf{case4\_ours} & 168 & 178.8 & 396 & 371.9 & 2.212 mW & -0.44 & -3.46 & 153 & 163.32 & 110.87 uW & 0.0 & 0.0 \\
\textbf{case5\_ours} & 10540 & 12016 & 6329 & 10642 & 170.7 mW & -4.14 & -409.54 & 5962 & 9600.21 & 19.98 mW & -1.02 & -56.72 \\
\textbf{case6\_ours} & 20 & 28.73 & 23 & 28.99 & 284.1 uW & -0.15 & -0.28 & 11 & 19.95 & 23.42 uW & 0.0 & 0.0 \\
\textbf{case7\_ours} & 23 & 35.64 & 23 & 34.58 & 385.6 uW & -0.14 & -0.47 & 14 & 27.40 & 35.59 uW & 0.0 & 0.0 \\
\textbf{case8\_ours} & 908 & 1063 & 2136 & 1916 & 14.75 mW & -0.69 & -44.12 & 709 & 939.78 & 746.77 uW & 0.0 & 0.0 \\
\textbf{case9\_ours} & 956 & 1173 & 2595 & 2354 & 19.54 mW & -0.95 & -61.83 & 847 & 1341.97 & 1.075 mW & 0.0 & 0.0 \\
\textbf{case10\_ours} & - & - & - & - & - & - & - & - & - & - & - &  \\
\textbf{case11\_ours} & 1 & 1.862 & 7 & 4.788 & 15.60 uW & 0.0 & 0.0 & 2 & 1.330 & 370.2 nW & 0.0 & 0.0 \\
\textbf{case12\_ours} & 1 & 1.062 & 1 & 1.064 & 2.588 uW & 0.0 & 0.0 & 1 & 1.064 & 258.8 nW & 0.0 & 0.0 \\
\textbf{case13\_ours} & - & - & - & - & - & - & - & - & - & - & - &  \\
\textbf{case14\_ours} & 4 & 1.862 & 6    & 7.98  & 28.15 uW & -0.1   & -0.1 & 5 & 5.054 & 1.767 uW & 0.0 & 0.0 \\\bottomrule
\end{tabular}
}
% \vspace{-.05in}
\caption{Reproduction PPA results of RTLRewriter optimized their own designs using Yosys and DC synthesis flows.}
\label{tab:exp8}
\end{table}

\section{Detailed LLM Evaluation Results and Statistical Analysis}
\label{appendix:llm-eval}

We \textbf{re-run each LLM evaluation 5 times} on RTL-OPT (36 designs), using the same prompt template to ensure consistency. For each run, we recorded the: 
(1) Syntax correctness; (2) Func correctness; (3) PPA better than suboptimal; (4) PPA better than optimized. The experimental results for 6 different LLMs (\textbf{including two open-source LLMs}) are summarized below.

We summarize statistical results and conduct paired t-tests on two representative LLM pairs: Gmini vs. DS-R1 and LLaMA vs. Qwen. We will include the complete \textbf{mean ± $\sigma$} and \textbf{paired t-tests} results in this appendix when updating the camera-ready version.

\subsection{Syntax Correctness (N out of 36)}
\resizebox{0.42\textwidth}{!}{
\begin{tabular}{c|c|c|c|c|c|c}
\toprule
Run & GPT-4o & Gmini-2.5 & DS V3 & DS R1 & QWEN & Llama \\
\midrule\midrule
1 & 36 & 35 & 36 & 32 & 34 & 15 \\
2 & 35 & 31 & 34 & 32 & 34 & 17 \\
3 & 36 & 30 & 36 & 31 & 31 & 15 \\
4 & 35 & 34 & 36 & 30 & 30 & 14 \\
5 & 35 & 35 & 36 & 32 & 34 & 15 \\
\midrule
\textbf{Mean} & 35.4 & 33 & 35.6 & 31.4 & 32.6 & 15.2 \\
\textbf{$\sigma$} & 1.0 & 1.5 & 1.3 & 0.8 & 0.9 & 1.1 \\
\bottomrule
\end{tabular}
}
\subsection{Functional Correctness (N out of 36)}
\resizebox{0.42\textwidth}{!}{
\begin{tabular}{c|c|c|c|c|c|c}
\toprule
Run & GPT-4o & Gmini-2.5 & DS V3 & DS R1 & QWEN & Llama \\
\midrule\midrule
1 & 29 & 26 & 27 & 24 & 23 & 13 \\
2 & 28 & 25 & 21 & 26 & 20 & 18 \\
3 & 25 & 24 & 26 & 25 & 23 & 17 \\
4 & 27 & 25 & 25 & 24 & 21 & 12 \\
5 & 28 & 28 & 25 & 26 & 19 & 17 \\
\midrule
\textbf{Mean} & 27.4 & 25.6 & 24.8 & 25 & 21.2 & 15.4 \\
\textbf{$\sigma$} & 1.5 & 1.5 & 2.3 & 1.8 & 1.6 & 2.7 \\
\bottomrule
\end{tabular}
}
\subsection{PPA \textgreater Suboptimal (N out of 36)}
\resizebox{0.42\textwidth}{!}{
\begin{tabular}{c|c|c|c|c|c|c}
\toprule
Run & GPT-4o & Gmini-2.5 & DS V3 & DS R1 & QWEN & Llama \\
\midrule\midrule
1 & 7 & 11 & 10 & 15 & 4 & 2 \\
2 & 8 & 9 & 6 & 16 & 4 & 5 \\
3 & 7 & 10 & 8 & 16 & 6 & 4 \\
4 & 7 & 9 & 10 & 13 & 4 & 1 \\
5 & 9 & 10 & 8 & 14 & 4 & 2 \\
\midrule
\textbf{Mean} & 7.6 & 9.8 & 8.4 & 14.8 & 4.4 & 2.8 \\
\textbf{$\sigma$} & 0.9 & 0.8 & 1.5 & 1.3 & 1.1 & 1.6 \\
\bottomrule
\end{tabular}
}
\subsection{PPA \textgreater Optimized (N out of 36)}
\resizebox{0.42\textwidth}{!}{
\begin{tabular}{c|c|c|c|c|c|c}
\toprule
Run & GPT-4o & Gmini-2.5 & DS V3 & DS R1 & QWEN & Llama \\
\midrule\midrule
1 & 2 & 2 & 5 & 5 & 1 & 0 \\
2 & 3 & 2 & 4 & 6 & 0 & 1 \\
3 & 1 & 4 & 5 & 5 & 2 & 1 \\
4 & 3 & 2 & 5 & 5 & 2 & 0 \\
5 & 2 & 2 & 3 & 4 & 1 & 1 \\
\midrule
\textbf{Mean} & 2.2 & 2.4 & 4.4 & 5 & 1.2 & 0.6 \\
\textbf{$\sigma$} & 0.8 & 0.9 & 1.0 & 0.7 & 0.8 & 0.5 \\
\bottomrule
\end{tabular}
}
\subsection{Comparison: GPT-4o vs. DS-R1}
\resizebox{0.6\textwidth}{!}{
\begin{tabular}{c|c|c|c|c|c}
\toprule
Metric & Mean Diff & t-value & p-value & Significance & Effect Size \\
\midrule\midrule
Syntax correctness & +3.8 & +6.53 & 0.0028 & Significant & 2.92 (Very Large) \\
Func correctness & +1.6 & +1.67 & 0.169 & Not Significant & 0.75 (Medium) \\
PPA \textgreater suboptimal & -7.2 & -5.81 & 0.004 & Significant & 2.60 (Very Large) \\
PPA \textgreater optimized & -2.8 & -5.83 & 0.004 & Significant & 2.61 (Very Large) \\
\bottomrule
\end{tabular}
}

\subsection{Comparison: LLaMA vs. Qwen}
\resizebox{0.6\textwidth}{!}{
\begin{tabular}{c|c|c|c|c|c}
\toprule
Metric & Mean Diff & t-value & p-value & Significance & Effect Size \\
\midrule\midrule
Syntax correctness & -18.2 & -48.40 & $<$0.0001 & Significant & -21.7 (Extreme) \\
Func correctness & -5.8 & -5.39 & 0.006 & Significant & -2.41 (Very Large) \\
PPA \textgreater suboptimal & +1.6 & +2.53 & 0.065 & Marginal & 0.72 (Medium) \\
PPA \textgreater optimized & +0.6 & +1.34 & 0.251 & Not Significant & 0.30 (Small) \\
\bottomrule
\end{tabular}
}

\iffalse
\section{Qualitative Analysis of LLM Functionality Failures}
\label{sec:app:quan}
As summarized in Section~\ref{sec:exp}, LLMs face a fundamental trade-off: conservative approaches avoid errors but miss optimization opportunities, while aggressive optimizations risk introducing functional bugs despite correct syntax.

To systematically analyze failure modes, we conducted a detailed study of 40 randomly sampled cases where LLM-generated optimizations passed syntax checks but introduced functional errors. \textbf{Our findings reveal three LLM failure categories}:

\begin{itemize} %[left=10pt]
    \item \textbf{Control Logic Inconsistencies} (19/40, 47.5\%)
    \begin{quote}
        Example (\texttt{comparator}): LLM failed to maintain bit-priority ordering (MSB-first comparison) and incorrectly implemented the ‘lt’ condition using wrong Boolean logic.
    \end{quote}
    
    \item \textbf{Overly Aggressive Pipelining} (12/40, 30\%)
    \begin{quote}
        Example (\texttt{fsm}): LLM reduced pipeline stages from 4 to 3 cycles, violating the original design’s latency requirements and causing incorrect output timing.
    \end{quote}
    
    \item \textbf{Improper Resource Sharing} (9/40, 22.5\%)
    \begin{quote}
        Example (\texttt{mux}): Register sharing ignored temporal dependencies, leading to stale data being read in subsequent cycles.
    \end{quote}
\end{itemize}

\fi
\newpage

% \section{LLM Prompt Generation Process}
% The complete implementation of the LLM prompt generation and evaluation pipeline for the RTL-OPT benchmark is available in the anonymous repository. The core prompt generation process is implemented in \texttt{Results/LLM\_Test\_result/llm\_gen.py}, (available in the anonymous repository).

%% file: iclr2026_conference.bib
@misc{designcompiler,
 title = {{Synoposys Design Compiler®}},
 howpublished = "\nolinkurl{https://www.synopsys.com/implementation-and-signoff/rtl-synthesis-test/design-compiler-nxt.html}",
    year = "2021",
 }

@misc{fomal,
 title = {{Formality® Equivalence Checking}},
 howpublished = "\nolinkurl{https://www.synopsys.com/implementation-and-signoff/signoff/formality-equivalence-checking.html}",
    year = "2023",
 }

@article{fang2025survey,
  title={A survey of circuit foundation model: Foundation ai models for vlsi circuit design and eda},
  author={Fang, Wenji and Wang, Jing and Lu, Yao and Liu, Shang and Wu, Yuchao and Ma, Yuzhe and Xie, Zhiyao},
  journal={arXiv preprint arXiv:2504.03711},
  year={2025}
}

@misc{xu2025rethinkingllmbasedrtlcode,
      title={Rethinking LLM-Based RTL Code Optimization Via Timing Logic Metamorphosis}, 
      author={Zhihao Xu and Bixin Li and Lulu Wang},
      year={2025},
      eprint={2507.16808},
      archivePrefix={arXiv},
      primaryClass={cs.SE},
      url={https://arxiv.org/abs/2507.16808}, 
}

@misc{vcs,
 title = {{VCS® functional verification solution}},
 howpublished = "\nolinkurl{https://www.synopsys.com/verification/simulation/vcs.html}",
    year = "2021",
 }

@article{liu2025deeprtl,
  title={DeepRTL: Bridging Verilog Understanding and Generation with a Unified Representation Model},
  author={Liu, Yi and Xu, Changran and Zhou, Yunhao and Li, Zeju and Xu, Qiang},
  journal={arXiv preprint arXiv:2502.15832},
  year={2025}
}

@article{wang2025symrtlo,
  title={SymRTLO: Enhancing RTL Code Optimization with LLMs and Neuron-Inspired Symbolic Reasoning},
  author={Wang, Yiting and Ye, Wanghao and Guo, Ping and He, Yexiao and Wang, Ziyao and Tian, Bowei and He, Shwai and Sun, Guoheng and Shen, Zheyu and Chen, Sihan and others},
  journal={arXiv preprint arXiv:2504.10369},
  year={2025}
}

@article{allam2024rtl,
 author = {Allam, Ahmed and Shalan, Mohamed},
 journal={ arXiv preprint arXiv:2405.17378 },
 title = {RTL-Repo: A Benchmark for Evaluating LLMs on Large-Scale RTL Design Projects},
 year = {2024}
}

@article{chang2023chipgpt,
 author = {Chang, Kaiyan and Wang, Ying and Ren, Haimeng and Wang, Mengdi and Liang, Shengwen and Han, Yinhe and Li, Huawei and Li, Xiaowei},
 journal={ arXiv preprint arXiv:2305.14019 },
 title = {ChipGPT: How far are we from natural language hardware design},
 year = {2023}
}

@article{cui2024origen,
 author = {Cui, Fan and Yin, Chenyang and Zhou, Kexing and Xiao, Youwei and Sun, Guangyu and Xu, Qiang and Guo, Qipeng and Song, Demin and Lin, Dahua and Zhang, Xingcheng and others},
 journal={ arXiv preprint arXiv:2407.16237 },
 title = {OriGen: Enhancing RTL Code Generation with Code-to-Code Augmentation and Self-Reflection},
 year = {2024}
}

@article{delorenzo2024creativeval,
 author = {DeLorenzo, Matthew and Gohil, Vasudev and Rajendran, Jeyavijayan},
 journal={ arXiv preprint arXiv:2404.08806 },
 title = {CreativEval: Evaluating Creativity of LLM-Based Hardware Code Generation},
 year = {2024}
}

@inproceedings{fu2023gpt4aigchip,
 author = {Fu, Yonggan and Zhang, Yongan and Yu, Zhongzhi and Li, Sixu and Ye, Zhifan and Li, Chaojian and Wan, Cheng and Lin, Yingyan Celine},
 booktitle={ International Conference on Computer-Aided Design (ICCAD) },
 title = {{GPT4AIGChip}: Towards next-generation {AI} accelerator design automation via large language models},
 year = {2023}
}

@article{ho2024verilogcoder,
 author = {Ho, Chia-Tung and Ren, Haoxing and Khailany, Brucek},
 journal={ arXiv preprint arXiv:2408.08927 },
 title = {Verilogcoder: Autonomous verilog coding agents with graph-based planning and abstract syntax tree (ast)-based waveform tracing tool},
 year = {2024}
}

@article{liu2023chipnemo,
 author = {Liu, Mingjie and Ene, Teodor-Dumitru and Kirby, Robert and Cheng, Chris and Pinckney, Nathaniel and Liang, Rongjian and Alben, Jonah and Anand, Himyanshu and Banerjee, Sanmitra and Bayraktaroglu, Ismet and others},
 journal={ arXiv preprint arXiv:2311.00176 },
 title = {{ChipNeMo: Domain-Adapted LLMs for Chip Design}},
 year = {2023}
}

@article{liu2023verilogeval,
 author = {Liu, Mingjie and Pinckney, Nathaniel and Khailany, Brucek and Ren, Haoxing},
 journal={ arXiv preprint arXiv:2309.07544 },
 title = {VerilogEval: Evaluating Large Language Models for Verilog Code Generation},
 year = {2023}
}

@article{liu2024craftrtl,
 author = {Liu, Mingjie and Tsai, Yun-Da and Zhou, Wenfei and Ren, Haoxing},
 journal={ arXiv preprint arXiv:2409.12993 },
 title = {CraftRTL: High-quality Synthetic Data Generation for Verilog Code Models with Correct-by-Construction Non-Textual Representations and Targeted Code Repair},
 year = {2024}
}

@article{liu2024rtlcoder,
 author = {Liu, Shang and Fang, Wenji and Lu, Yao and Zhang, Qijun and Zhang, Hongce and Xie, Zhiyao},
 journal={ IEEE Transactions on Computer-Aided Design of Integrated Circuits and Systems (TCAD) },
 title = {RTLCoder: Fully Open-Source and Efficient LLM-Assisted RTL Code Generation Technique},
 year = {2024}
}

@inproceedings{lu2024rtllm,
 author = {Lu, Yao and Liu, Shang and Zhang, Qijun and Xie, Zhiyao},
 booktitle={ Asia and South Pacific Design Automation Conference (ASP-DAC) },
 title = {{RTLLM}: An Open-Source Benchmark for Design RTL Generation with Large Language Model},
 year = {2024}
}

@misc{NanGate,
 author = {Si2},
 title = {{NanGate} 45nm Open Cell Library},
 year = {2018}
}

@article{pei2024betterv,
 author = {Pei, Zehua and Zhen, Hui-Ling and Yuan, Mingxuan and Huang, Yu and Yu, Bei},
 journal={ arXiv preprint arXiv:2402.03375 },
 title = {BetterV: Controlled Verilog Generation with Discriminative Guidance},
 year = {2024}
}

@article{pinckney2024revisiting,
 author = {Pinckney, Nathaniel and Batten, Christopher and Liu, Mingjie and Ren, Haoxing and Khailany, Brucek},
 journal={ arXiv preprint arXiv:2408.11053 },
 title = {Revisiting VerilogEval: Newer LLMs, In-Context Learning, and Specification-to-RTL Tasks},
 year = {2024}
}

@article{thakur2023autochip,
 author = {Thakur, Shailja and Blocklove, Jason and Pearce, Hammond and Tan, Benjamin and Garg, Siddharth and Karri, Ramesh},
 journal={ arXiv preprint arXiv:2311.04887 },
 title = {AutoChip: Automating HDL Generation Using LLM Feedback},
 year = {2023}
}

@inproceedings{wolf2013yosys,
 author = {Wolf, Clifford and Glaser, Johann and Kepler, Johannes},
 booktitle={ Austrian Workshop on Microelectronics (Austrochip) },
 title = {Yosys-a free Verilog synthesis suite},
 year = {2013}
}

@article{yao2024rtlrewriter,
 author = {Yao, Xufeng and Wang, Yiwen and Li, Xing and Lian, Yingzhao and Chen, Ran and Chen, Lei and Yuan, Mingxuan and Xu, Hong and Yu, Bei},
 journal={ arXiv preprint arXiv:2409.11414 },
 title = {RTLRewriter: Methodologies for Large Models aided RTL Code Optimization},
 year = {2024}
}

@article{zhao2024codev,
 author = {Zhao, Yang and Huang, Di and Li, Chongxiao and Jin, Pengwei and Nan, Ziyuan and Ma, Tianyun and Qi, Lei and Pan, Yansong and Zhang, Zhenxing and Zhang, Rui and others},
 journal={ arXiv preprint arXiv:2407.10424 },
 title = {CodeV: Empowering LLMs for Verilog Generation through Multi-Level Summarization},
 year = {2024}
}

@inproceedings{liu2024openllm,
 author = {Liu, Shang and Lu, Yao and Fang, Wenji and Li, Mengming and Xie, Zhiyao},
 booktitle={ International Conference on Computer-Aided Design (ICCAD) },
 title = {OpenLLM-RTL: Open Dataset and Benchmark for LLM-Aided Design RTL Generation},
 year = {2024}
}

@article{pinckney2025comprehensive,
  title={Comprehensive Verilog Design Problems: A Next-Generation Benchmark Dataset for Evaluating Large Language Models and Agents on RTL Design and Verification},
  author={Pinckney, Nathaniel and Deng, Chenhui and Ho, Chia-Tung and Tsai, Yun-Da and Liu, Mingjie and Zhou, Wenfei and Khailany, Brucek and Ren, Haoxing},
  journal={arXiv preprint arXiv:2506.14074},
  year={2025}
}
